# Slum Detection and Density Mapping with AlphaEarth Foundations: A Representation Learning Evaluation Across 12 Global Cities

Shuyang Hou[a], Ziqi Liu[a], Haoyue Jiao[a], Zhangyan Xu[a], Xiaopu Zhang[a], Lutong Xie[a], Yaxian Qing[a]*, Jianyuan Liang[a], Xuefeng Guan[a], Huayi Wu[a]

*a. State Key Laboratory of Information Engineering in Surveying, Mapping, and Remote Sensing, Wuhan University, Wuhan, China

*Corresponding author: Yaxian Qing, email: qingyaxian@whu.edu.cn

## Abstract

Pixel-level slum mapping has long been constrained by limited cross-city generalisation, the absence of continuous density estimation, and weak global comparability. AlphaEarth Foundations (AEF), a globally consistent 64-dimensional annual surface embedding at 10 m, offers a new analysis-ready basis for lightweight slum monitoring, but its applicability to slum detection — an indirectly coupled task shaped by both built form and socio-economic processes — remains untested. We evaluate AEF on slum classification and sub-pixel density estimation across 12 cities and 69 city–year pairs (2017–2024), using GRAM pseudo-masks as supervisory labels. The evaluation spans four training strategies, two protocols (random split and 3×3 spatial block cross-validation), six auxiliary feature configurations, and five baseline models, complemented by representation-level analyses (PCA, SHAP) and full-AOI mapping. Five findings emerge. (1) Same-city cross-year training is optimal under both protocols (median spatial F1 = 0.616, $R^2$ = 0.466); temporal expansion outperforms cross-city transfer, indicating city-scale representational drift. (2) Regression $R^2$ is driven primarily by zero/non-zero boundary discrimination: positive-pixel $R^2$ is consistently negative across all cities, revealing limited capacity to model intra-pixel density gradients at 10 m. (3) PC36 is consistently top-ranked across tasks; classification saturates at k = 32 while regression remains unsaturated at k = 64. (4) POI features yield the largest density gain ($\Delta R^2$ = +0.064). (5) For six cities meeting dual-task usability thresholds, full-AOI inference across 2017–2024 preserves slum cluster structure (mean SSIM = 0.926). The study delineates the capabilities and complementarity needs of foundation-model embeddings for slum monitoring.

## 1. Introduction

Slums are informal settlements characterized by the confluence of housing shortages, inadequate infrastructure, and pronounced spatial inequality under conditions of rapid urbanization(Chi et al., 2022; Rolf et al., 2021). According to estimates by UN-Habitat, approximately one billion people worldwide — roughly 30% of the urban population — reside in such settlements(Aerni, 2016). The spatial boundaries and internal densities of slums critically influence the allocation of public resources, including water supply, sanitation, healthcare, and disaster risk mitigation(Saad, 2021). However, most related data rely on field surveys, population censuses, or administrative records, which are constrained by cost, update frequency, definitional criteria, and institutional contexts, thereby limiting the availability of globally comparable spatial information(Montana et al., 2016; Pedro and Queiroz, 2019). Developing a timely, standardized, and globally comparable spatial data framework for slums remains a central challenge for urban poverty monitoring and policy formulation(Kuffer et al., 2016; Kuffer et al., 2018; Verma et al., 2019).

Remote sensing provides the observational foundation for large-scale and repeatable mapping of slums(Kuffer et al., 2016). Early studies employed medium-to high-resolution optical imagery combined with handcrafted features — such as rooftop reflectance, building density, texture, and road networks — together with classifiers like SVM or RF to perform binary classification(Jean et al., 2016; Kit et al., 2012; Taubenböck and Kraff, 2014).

Deep learning approaches have substantially improved detection accuracy(Raj et al., 2024; Yeh et al., 2020); however, evaluations often rely on random data splits, which fail to reflect generalization performance under spatially independent conditions(Büttner et al., 2025; Verma et al., 2019). GRAM achieved an overall mean Intersection-over-Union (mIoU) of 0.86 on three target cities in Africa using ESRI very-high-resolution (VHR, ~1.2 m) imagery, through expert mixing and test-time adaptation, representing the current state of the art(Lee et al.). Nonetheless, it depends on VHR imagery (<1 m), incurs high computational costs, and its global scalability is limited. In terms of density modeling, existing research largely remains at the district-level poverty or wealth index regression(Patel et al., 2020), where the target variables are aggregate socioeconomic indicators rather than physical density(Haregu et al., 2018). Pixel-scale modeling of sub-pixel slum density remains comparatively underdeveloped.

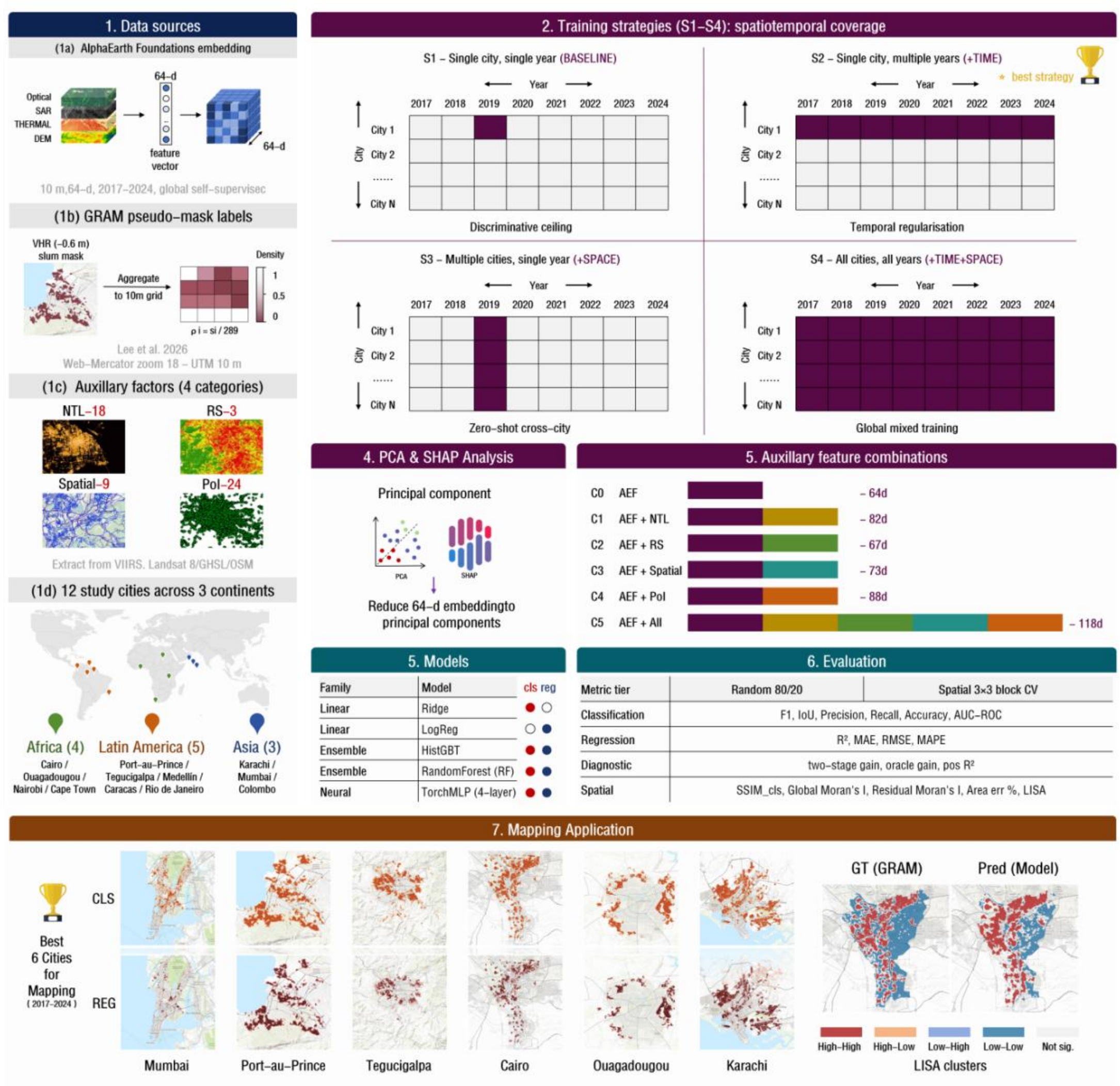

**Fig. 1 Overview of the Research Framework.**

Released in July 2025, AlphaEarth Foundations (AEF, Google DeepMind) offers a novel approach to addressing the aforementioned challenges(Brown et al., 2025). AEF integrates multi-source observations—including optical imagery, SAR, climate data, and textual information—at approximately 10 m resolution on a global scale, generating 64-dimensional annual surface embeddings through self-supervised learning, and publishes them as "analysis-ready data products." Compared with earlier foundational models such as Tile2Vec(Jean et al.), SeCo(Yao et al.), and Tessera(Feng et al., 2025), AEF provides globally consistent coverage, annual updates, automatic multi-source fusion, and embeddings that can be directly applied to downstream modeling tasks. Existing studies have demonstrated the effectiveness of these representations in agricultural mapping(Ma et al., 2026; Wang et al., 2026; Yang et al., 2026), functional zone classification(Benavides-Martinez et al., 2026; Liu et al., 2025), and land cover mapping(Xiao et al., 2026). However, slums are influenced simultaneously by built environment morphology and socioeconomic processes, making them an indirect mapping problem; their spectral and textural patterns vary considerably across cities and climatic conditions. Whether the 64-dimensional AEF embeddings can learn stable and discriminative representations under such heterogeneity remains to be systematically evaluated.

Building on the perspectives outlined above, this study addresses five core questions:

- Q1 (Feasibility): Can the 64-dimensional AEF embeddings support both slum classification and density estimation under four data organization strategies (S1–S4)?
- Q2 (Dimensional Structure): Are the key embedding dimensions consistent across the two tasks? Do stable backbone dimensions exist across tasks and cities?
- Q3 (Auxiliary Gains): What is the marginal contribution of four types of auxiliary factors—nighttime lights, remote sensing indices, spatial structure, and points of interest (POIs)? Are the optimal configurations consistent for both classification and density estimation?
- Q4 (Full-Scene Inference): When applied to the full extent of AEF imagery, including years without annotations, are the model predictions visually coherent?
- Q5 (Spatial Structure): Beyond pixel-level accuracy, do the predictions capture the overall spatial clustering patterns of slums?

Building on the five core questions outlined above, this study establishes a systematic evaluation framework encompassing twelve cities, dual tasks, five models, four training strategies, and five types of auxiliary configurations. Slum sub-pixel density regression is assessed at the pixel scale, quantifying performance differences between spatially independent and random data splits. From a representation learning perspective, the study characterizes the dimensional structure of AEF embeddings and the task-specific contributions of auxiliary factors. For the subset of cities meeting dual-task usability thresholds, full-scene inference and inter-annual prediction interpolation from 2017 to 2024 are performed. The structural fidelity of these predictions is validated using spatial statistics, including SSIM, Moran's I, and LISA, as illustrated in Figure 1.

The contributions of this study are as follows:

- This is the first systematic evaluation of AEF global surface embeddings for dual tasks of slum detection and density estimation, covering twelve representative cities worldwide.
- A pixel-scale evaluation paradigm for sub-pixel slum density estimation is established, addressing the methodological gap in existing studies that largely focus on district-level aggregate indicators.
- The study quantifies the generalization decay of foundational model embeddings under strictly spatially independent splits, providing a methodological reference for cross-domain evaluation of remote sensing foundational models.
- From a representation learning perspective, the study reveals the systematic division of labor between AEF embedding dimensions and auxiliary factors across classification and density tasks.
- Full-scene inference and inter-annual prediction interpolation from 2017 to 2024 are conducted for six cities meeting dual-task usability thresholds, demonstrating the feasibility of applying foundational model embeddings to practical mapping applications.

The remainder of this paper is organized as follows: Chapter 2 introduces the study areas, the construction of AEF embeddings, GRAM pseudo-mask labels, and auxiliary factors. Chapter 3 presents the benchmark models, training strategies, and the evaluation metric system. Chapter 4 reports the direct performance of the dual tasks—classification and density estimation—as well as the analysis of embedding dimensional structure and auxiliary factor contributions, corresponding to Q1–Q3. Chapter 5 conducts full-scene inference and spatial structure validation for the subset of cities meeting dual-task usability thresholds, addressing Q4–Q5. Chapter 6 discusses the main findings, situates the work within existing literature, and identifies research limitations. Finally, Chapter 7 concludes the study and outlines directions for future work.

## 2. Study Areas and Data

### 2.1. Study Areas

This study selects twelve representative cities worldwide (Table A1), covering three major regions: Africa (4 cities), Latin America (5 cities), and Asia (3 cities). City selection follows three criteria:(1) Availability of GRAM pixel-level annotations with reliable coverage that spatially aligns with AEF embeddings.(2) Geographic and urban morphological diversity, encompassing a variety of spatial patterns (see the "Spatial Pattern" column in Table A1).(3)A wide gradient of slum pixel proportions within the study areas, ranging from 1.4% in Colombo to 22.6% in Karachi (Table A3), enabling evaluation of model performance under different class balance conditions. The temporal coverage of each city is discontinuous, constrained by the availability of VHR imagery in the GRAM source dataset (see Table A2). The experimental corpus comprises 69 city–year pairs spanning 2017–2024. The twelve selected cities fully correspond to the twelve source-domain cities in the GRAM training set(Lee et al.), ensuring that pseudo-mask labels retain the highest quality within the original training data distribution.

### 2.2. Data and Preprocessing

#### 2.2.1. AEF

AEF data for the twelve cities from 2017 to 2024 were downloaded via the GEE platform, comprising a total of 136 GeoTIFF files (~85 GB). For each city, data are stored in 64-band UTM format with a spatial resolution of approximately 10 m, and individual file sizes range from 0.72 to 1.9 GB.

#### 2.2.2. GRAM Pseudo-Mask Annotations

GRAM (Lee et al., 2026) represents the currently most widely cross-continental pixel-level slum segmentation framework. It combines a Mixture-of-Experts approach with Test-Time Adaptation and uses ESRI World Imagery (zoom=16, ~1.2 m) as input. On three target cities in Africa, it achieves an overall mIoU of 0.859 (F1 = 0.921), with the IoU in Dar es Salaam improved from the baseline 0.659 to 0.752. The pseudo-mask predictions are publicly released at higher resolution (zoom=18 tiles, 256 × 256 pixels, ~0.6 m, EPSG:3857), providing wide coverage, fine granularity, and annual updates. These predictions have been validated against manual annotations in the original publication and can serve as supervisory signals for remote sensing studies of slums.

In this study, the zoom=18 pseudo-mask tiles published by GRAM are directly used to construct supervisory labels. Using each city's AEF UTM coordinate system and 10 m grid as the reference, GRAM tiles are batch-transformed to the corresponding UTM system to establish coverage mappings from 0.6 m sub-pixels to 10 m pixels. Each i-th 10 m pixel contains $N_i = 17 \times 17 = 289$sub-pixels of 0.6 m. Let $s_i$denote the number of slum sub-pixels; the sub-pixel slum density is then defined as:

$$\rho_i = \frac{s_i}{N_i} \in [0,1] \tag{1}$$

Based on this, dual-task supervisory labels are constructed: the classification label $y_i^{\text{cls}} = 1$if $s_i > 0$(i.e., any sub-pixel is predicted as slum, consistent with the GRAM "presence implies presence" semantics); the regression label $y_i^{\text{reg}} = s_i \in \{0,1, ..., N_i\}$serves as a continuous target for sub-pixel decomposition. AEF NoData values (-9999.0) are uniformly filtered; in regions of overlapping tiles, the classification label takes the maximum value, and the regression label takes the maximum density.

Diagnostics across all 383 million sample pixels show that $\rho_i$ exhibits a strongly bimodal distribution: 89.3% of pixels have $s_i = 0$ (completely slum-free), 8.6% have $\rho_i > 0.9$, and the intermediate range ($0 < \rho_i \leq 0.9$) accounts for only 2.1%. The degree of bimodality varies significantly across cities (Figure 2): the proportion of zero-density pixels increases from Karachi (77.3%), Port-au-Prince (77.7%), and Cairo (79.9%) to Cape Town (98.5%) and Colombo (98.7%), spanning over 21 percentage points. In most cities, the median density of non-zero pixels is close to 1.0, indicating that once a pixel is classified as slum, almost all sub-pixels are slum-covered. The only exception is Tegucigalpa (median ~0.75), where the intermediate transition zone is thicker, reflecting blurred boundaries between slums and formal built-up areas in mountainous terrain.

This distribution indicates that at 10 m resolution, slum boundaries are spatially well-defined: the classification captures actual boundaries rather than artificial thresholds. The added value of the regression task is concentrated in the 2.1% of pixels in boundary transition zones, characterizing density heterogeneity that cannot be distinguished by classification alone.

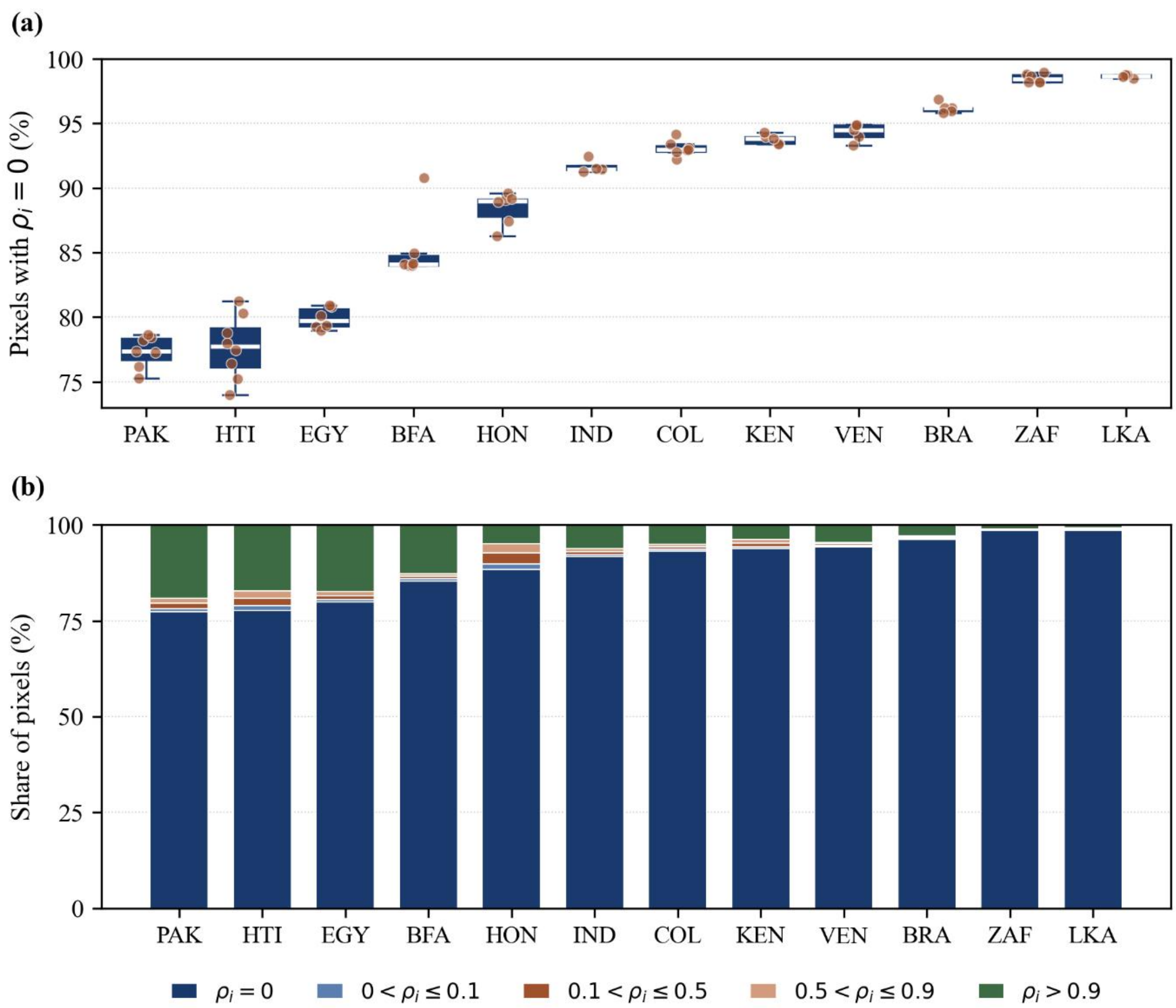

**Fig. 2 Cross-city heterogeneity of slum density label distributions across 69 city–year pairs.** (a) Per-city distribution of $\rho_i = 0$ share; box plots show inter-year variation, dots are individual years. Cities are ordered by ascending $\rho_i = 0$ median, i.e., from the most slum-dense (PAK) to the least (LKA). (b) Pixel share in five $\rho_i$ bins, averaged across years per city; intermediate bins (light blue, red-orange, light orange) appear visually thin due to the bimodal distribution. City order matches (a).

### 2.2.3. Auxiliary Factors

In addition to the 64-dimensional AEF baseline, four types of auxiliary factors are introduced to supplement semantic signals not captured by AEF visual embeddings. These factors span four dimensions: nighttime lights, biophysical indices, spatial structure, and social services. Detailed data sources, native resolutions, and feature compositions are summarized in Table 1.

**Table 1. Auxiliary feature categories: data sources, native resolution, content, and inter-city heterogeneity.** Cross-city CV is the standard deviation of city-level means divided by the global mean; lower values indicate more stable features across cities.

| Category | Dim | Source | Native resolution | Content | Mean CV | Max CV (feature) |
|---|---|---|---|---|---|---|
| **NTL** | 18 | VIIRS DNB monthly composite + DMSP-OLS 2013 baseline | ~500 m → 10 m | Annual statistics (mean / median / max / min / SD / P10 / P90 / CV / lit ratio); seasonal difference; 3×3 focal mean & contrast; stray light; growth since 2013 | 0.74 | 3.01 (NTL_growth_2013) |
| **RS** | 3 | Landsat 8 annual median; JRC GHSL | 10—30 m → 10 m | NDVI; NDBI; built-up density (interpolated from 2015 & 2020 GHSL) | 0.61 | 1.23 |
| **Spatial** | 9 | OSM annual snapshot | Vector → 10 m | Distance to city centre / water body; road density (motorway-trunk-primary, secondary-tertiary, residential-service, total); distance to subway; subway count within 1 km; distance to bus stop | **0.48** | 1.09 |
| **POI** | 24 | OSM annual snapshot | Vector → 10 m | 12 facility classes (food, education, health, finance, retail, leisure, public service, recreation, accommodation, religion, transit, office); kernel density + nearest-distance per class | 0.74 | 2.88 (poi_leisure_density) |

Let the AEF embedding for the i-th pixel be $a_i \in \mathbb{R}^{64}$. The full feature vector is defined as:

$$x_i^{(k)} = \left[a_i \ \| \ f_i^{\mathrm{NTL}} \ \| \ f_i^{\mathrm{RS}} \ \| \ f_i^{\mathrm{Spatial}} \ \| \ f_i^{\mathrm{POI}}\right] \tag{2}$$

where $f_i^{\mathrm{NTL}} \in \mathbb{R}^{18}$, $f_i^{\mathrm{RS}} \in \mathbb{R}^{3}$, $f_i^{\mathrm{Spatial}} \in \mathbb{R}^{9}$, and $f_i^{\mathrm{POI}} \in \mathbb{R}^{24}$. Inclusion of each component is determined by the combination index $k$. This study defines six combinations (C0–C5), ranging from the pure AEF baseline (C0) to full-factor integration (C5), as detailed in Table 2.

**Table 2. Feature combination codes.** Each configuration concatenates the AEF 64-dim baseline with selected auxiliary categories. C0 serves as the pure-AEF baseline; C1 – C4 isolate the marginal contribution of each individual category; C5 stacks all four.

| Code | Composition | Total dim |
|---|---|---|
| C0 | AEF (baseline) | 64 |
| C1 | AEF + NTL | 82 |
| C2 | AEF + RS | 67 |
| C3 | AEF + Spatial | 73 |
| C4 | AEF + POI | 88 |
| C5 | AEF + NTL + RS + Spatial + POI | 118 |

Auxiliary factors differ substantially in numerical range and scale (e.g., NDVI $\in [-1,1]$ versus POI kernel density spanning several orders of magnitude). To prevent high-dimensional features from dominating the optimization of linear models, linearly sensitive baselines employ a RobustScaler based on the median and interquartile range of the training set to suppress extreme values. Tree-based ensemble models, being insensitive to monotonic transformations, use the raw feature scales directly. Deep learning models perform implicit layer-wise normalization via BatchNorm layers (see section 3.1). AEF embeddings are globally normalized by Google Earth Engine, and all models directly use their original values as input.

Cross-city coefficient of variation (CV across cities) analysis (the last two columns of Table 1) indicates that the stability of auxiliary factors increases in the order: Spatial (mean CV = 0.48) < RS (0.61) < NTL (0.74) ≈ POI (0.74). Among individual features, NTL_growth_2013 (CV = 3.01), poi_leisure_density (CV = 2.88), and poi_education_density (CV = 1.61) exhibit the strongest heterogeneity, reflecting systematic differences in urban development trajectories and OSM coverage. Such variability may affect cross-city transfer performance (see section 4.1).

## 3. Methods

### 3.1. Models

This study employs three types of models—linear models, ensemble learning models, and neural networks—comprising a total of five benchmark models. Detailed configurations are provided in Table 3. AEF embeddings are fixed per-pixel feature vectors, naturally compatible with tabular input formats. The core objective of this study is to evaluate the discriminative power of AEF representations themselves; therefore, lightweight models with simple structures and few hyperparameters are selected to avoid additional confounding effects introduced by complex model architectures.

**Table 3. Baseline models and hyperparameter configurations used for slum classification and density regression.**

| Task | Model | Hyperparameters |
|---|---|---|
| Classification | LogReg | L2 regularisation, C = 1.0, max_iter = 1000 |
| | HistGBT | max_depth = 6, max_iter = 200, learning_rate = 0.1 |
| | RF | n_estimators = 200, max_depth = 12, min_samples_leaf = 5 |
| | TorchMLP | 4-layer MLP, hidden 512→256→128→64; BCEWithLogitsLoss + pos_weight; Adam, lr = 1e-3; early stop patience = 20 |
| Regression | Ridge | α = 1.0 |
| | HistGBT | max_depth = 6, max_iter = 200, learning_rate = 0.1; HuberLoss, δ = 10.0 |
| | RF | n_estimators = 200, max_depth = 12, min_samples_leaf = 5 |
| | TorchMLP | Same architecture as classification version; HuberLoss, δ = 10.0 |

### 3.2. Experimental Design

The overall experimental design is organized into three hierarchical layers. The first layer (sections 3.2.1) compares four data organization strategies (S1–S4) using the pure AEF embeddings to address Q1 and to select representative strategies. The second layer (sections 3.2.2–3.2.3) analyzes embedding dimensional structure (Q2) and auxiliary factor contributions (Q3) under the selected strategies. The third layer (sections 3.2.4–3.2.5) evaluates the scalability of practical mapping applications (Q4) and the structural fidelity of spatial predictions (Q5).

#### 3.2.1. Direct Training Strategies (S1–S4)

Using the 64-dimensional AEF embeddings as input and GRAM pseudo-masks as supervisory signals, no auxiliary factors are added, and no dimensionality reduction is performed. The influence of training set spatiotemporal coverage on model generalization is systematically evaluated. The four strategies incrementally expand the training scope; definitions are detailed in Table 4.

**Table 4. Four direct training strategies for evaluating the spatiotemporal coverage of training data.**

| Strategy | Training data | Test data | Purpose |
|---|---|---|---|
| S1 | Target city, target year | Target city, target year | Baseline: single-city single-year discriminative ceiling |
| S2 | Target city, all available years | Target city, target year | Temporal regularisation effect |
| S3 | Other cities, same year | Target city, target year | Zero-shot cross-city spatial generalisation |
| S4 | All cities and years (excluding target city–year pair) | Target city, target year | Global mixed training, generalisation upper bound |

For each strategy, metrics are computed under both random (80/20 split) and spatial (3×3 spatial block cross-validation) evaluation schemes. For S2–S4, training set sizes are aligned with the target city–year training complement (~480,000 pixels) via proportional sampling based on each source city or year, ensuring that performance differences among S1–S4 arise from spatiotemporal composition rather than sample size. The strategy exhibiting the best performance under both random and spatial evaluations is selected as the baseline for subsequent experiments, on which dimensional structure and auxiliary factor effects are further explored.

#### 3.2.2. Embedding Dimensional Structure Analysis

The internal structure of the 64-dimensional AEF embeddings is assessed from two perspectives:

- PCA Ablation: AEF embeddings are compressed to $k \in \{8,16,24,32,38,48,56,64\}$ principal components via PCA. Under the S1 baseline strategy, all 69 city–year pairs are retrained, and performance of classification and regression tasks is observed under both random and spatial evaluation schemes. Cumulative explained variance curves are used to quantify the dimensional requirements of AEF embeddings from the perspectives of task type and evaluation scheme.
- SHAP Importance Analysis: Under the selected optimal strategy, each city–year pair is independently trained, and the mean absolute SHAP values for each AEF dimension are computed. Linear models use LinearSHAP, tree-based models use TreeSHAP, and TorchMLP uses gradient-based SHAP. The consistency of cross-city dimension importance rankings is evaluated via 1,000 permutation tests for statistical significance.

### 3.2.3. Auxiliary Factor Integration

On top of the 64-dimensional AEF baseline (C0), four types of auxiliary factors are concatenated to form combinations C1–C5 (see Table 2). Under the selected optimal strategy, models are trained and evaluated independently for each combination. Marginal gains are computed relative to C0 to distinguish discriminative improvements due to auxiliary factors from threshold optimization effects arising from probability distribution shifts.

### 3.2.4. Full-Scene Inference and Inter-Annual Applications

To assess the scalability of the framework in practical mapping tasks (Q4), for the subset of cities meeting dual-task usability thresholds (see section 4.3), a single TorchMLP is trained for each city using samples from all available GT years and the city-specific optimal auxiliary configuration. The trained models are then applied to the full AEF imagery for all eight years from 2017 to 2024 (each city covering 100–2,000 km², with 1–20 million effective pixels; see Table A3), enabling both accuracy reproduction for annotated years and prediction interpolation for unannotated years.

### 3.2.5. Spatial Structural Fidelity Validation

For the full-scene inference results (section 3.2.4), spatial structure metrics (see section 3.3.3) are computed on the subset of GT-annotated years. Local Indicators of Spatial Association (LISA) four-quadrant classification (HH/LL/HL/LH/NS) is used as a visual validation of spatial clustering fidelity. Note that GT labels are GRAM pseudo-masks; thus, validation should be interpreted as the structural consistency between model predictions and GRAM labels rather than absolute ground truth.

## 3.3. Evaluation Metrics

### 3.3.1. Pixel-Level Accuracy Metrics

For the classification task, the primary metrics are F1 score and IoU (Jaccard index), calculated specifically for the positive class (slum pixels). Due to class imbalance, metrics for the negative class are consistently close to 1 across cities and models and therefore are not informative. Auxiliary metrics include Precision, Recall, Accuracy, and AUC-ROC. Under the spatial cross-validation scheme, the standard deviation of F1 scores across folds ($F1_{std}$) is additionally recorded to assess stability across spatial blocks.

For the regression task, the primary metrics are $R^2$ and MAE (units: number of slum sub-pixels). Auxiliary metrics include RMSE and MAPE; the latter is computed only on the subset of pixels with $y_i > 0$, as the systematic scale difference between 10 m AEF and 0.6 m GRAM data introduces substantial variance and thus MAPE is not used as a primary metric.

All cross-sample aggregated metric values are computed using a fold-median approach: for each (city, year) sample, median values are first calculated across evaluation folds, and then summarized across samples. Additionally, in section 4.1, models are ranked per (city, year) sample according to F1/R², and cross-sample average ranks and win counts (number of times a model ranks first) are reported to assess the relative stability of model advantages.

### 3.3.2. Decomposition Metrics for Regression $R^2$

Given the strongly bimodal distribution of density labels (section 2.2.2 shows 89.3% of pixels are zero, 8.6% in the high-density tail), the interpretation of single-stage $R^2$ can be dominated by correctly predicted zero pixels, potentially masking true fit performance on positive pixels. To support the decomposition analysis in section 4.1.4, three diagnostic metrics are defined:

- **Two-stage gain:** $R^2$ recalculated after forcing pixels predicted as zero by the classifier to zero, quantifying the marginal improvement of a hard-cascaded two-stage approach relative to single-stage regression.

- **Oracle gain:** $R^2$ computed by replacing the classifier prediction with the true $y_{\text{cls}}$ for zeroing, representing the upper bound under an ideal classifier.
- **Positive $R^2$ (pos $R^2$):** $R^2$ computed only on the subset of positive pixels ($y_i > 0$), isolating the contribution of correct zero-pixel predictions to the primary $R^2$ metric.

#### 3.3.3. Spatial Structure Metrics

For full-scene inference results (section 3.2.4), three categories of spatial structure metrics are introduced to quantify structural consistency between predictions and GRAM labels:

- **Morphological similarity:** $\text{SSIM}_{\text{cls}}$ quantifies similarity in cluster shape, boundaries, and internal texture between predicted and ground truth binary classification.
- **Spatial clustering intensity and residual structure:** Global Moran's I is computed separately on GT and Pred to quantify whether the model preserves slum spatial clustering. Residual Moran's I measures whether prediction errors exhibit spatial structure (significantly positive values indicate error clustering in specific subregions rather than random distribution).
- **Aggregate area deviation:** $\text{Area_pct_err} = |\text{Pred area} - \text{GT area}| / \text{GT area} \times 100\%$ , serving as an application-oriented aggregate metric.

Additionally, Local Indicators of Spatial Association (LISA; Anselin, 1995) four-quadrant classification (High–High / Low–Low / High–Low / Low–High / Not Significant) is used to visually compare pixel-wise spatial clustering patterns between predictions and GT, providing visual evidence of spatial clustering fidelity. Moran's I and LISA are computed on adaptively downsampled grids (4×–8×) with ≤600,000 cells for Moran's I and ≤250,000 cells for LISA, using Queen contiguity (8 neighbors, row-standardized), with 99 permutations and a significance threshold of $p < 0.05$.

## 4. Results

### 4.1. Direct Application Performance

#### 4.1.1. Classification Task Performance

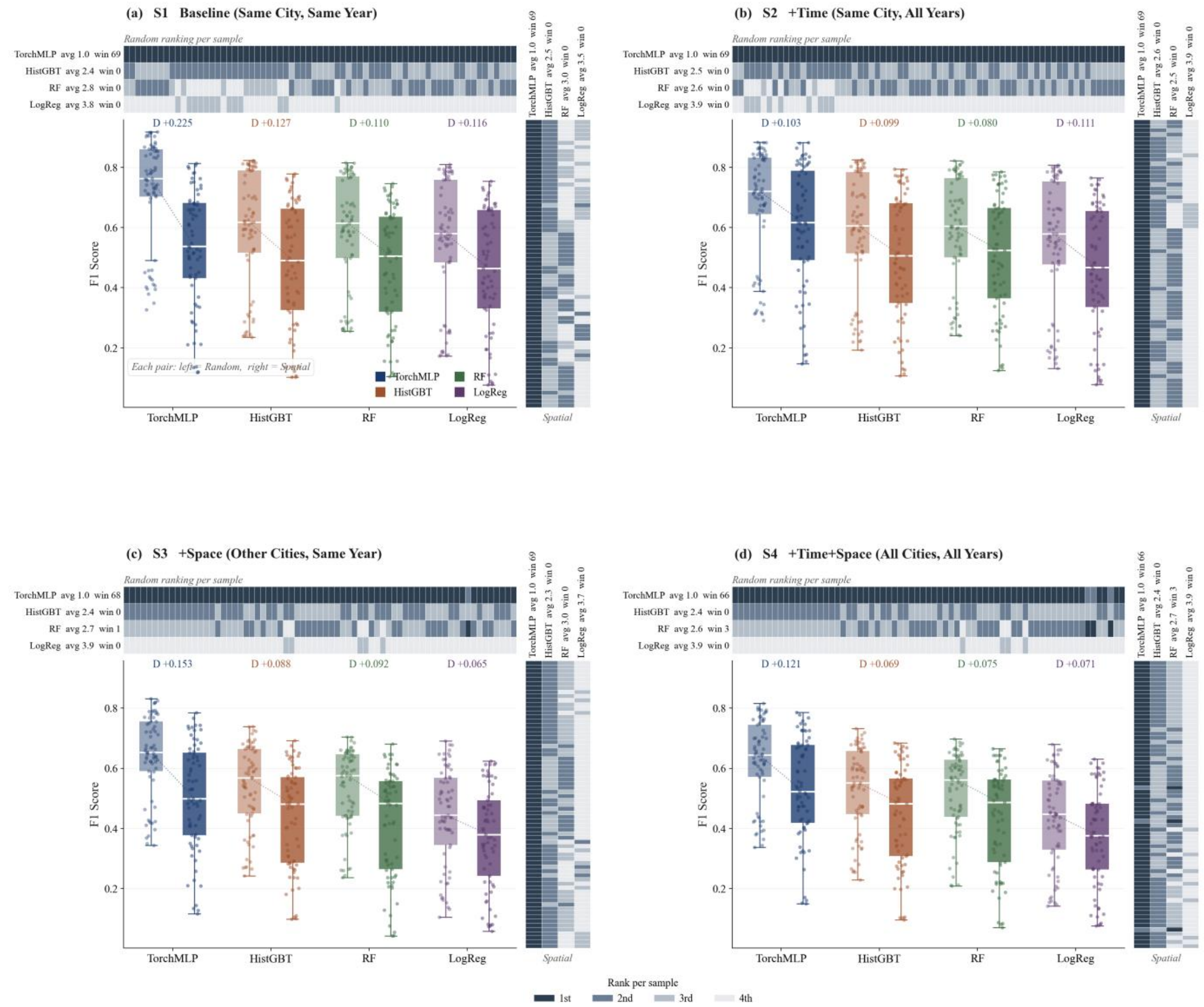

**Fig. 3 Classification F1 across four training strategies (panels a – d: S1 – S4) and four classifiers (LogReg, HistGBT, RF, TorchMLP).** Box plots pair Random (left) and Spatial (right) F1 over 69 city – year pairs, with median Δ annotated above each pair. Top and right strips show per-sample model rankings under Random and Spatial evaluation; rank colours range from 1st (darkest) to 4th (lightest), with each model's average rank and win count labelled.

As shown in Figure 3, TorchMLP consistently achieves the highest median F1 scores under random evaluation across all four strategies (S1: 0.763; S2: 0.720), with an average rank close to 1.0. However, it also exhibits the largest drop from random to spatial evaluation (S1: Δ = 0.225). In comparison, HistGBT, RF, and LogReg under S1 have Δ values of 0.127, 0.110, and 0.116, respectively, all significantly lower than TorchMLP, reflecting that tree-based ensemble and linear models are less sensitive to spatial distribution shifts. LogReg performs the weakest across all strategies (median spatial F1 = 0.38–0.46), consistently ranking fourth across samples, indicating that the relationship between AEF embeddings and slum labels cannot be adequately captured by a linear model. Model rankings are highly consistent between random and spatial evaluations (top/right strips in Figure 3): TorchMLP always ranks first, LogReg always last, with the random→spatial gap mainly reflecting overall performance decline rather than rank reshuffling.

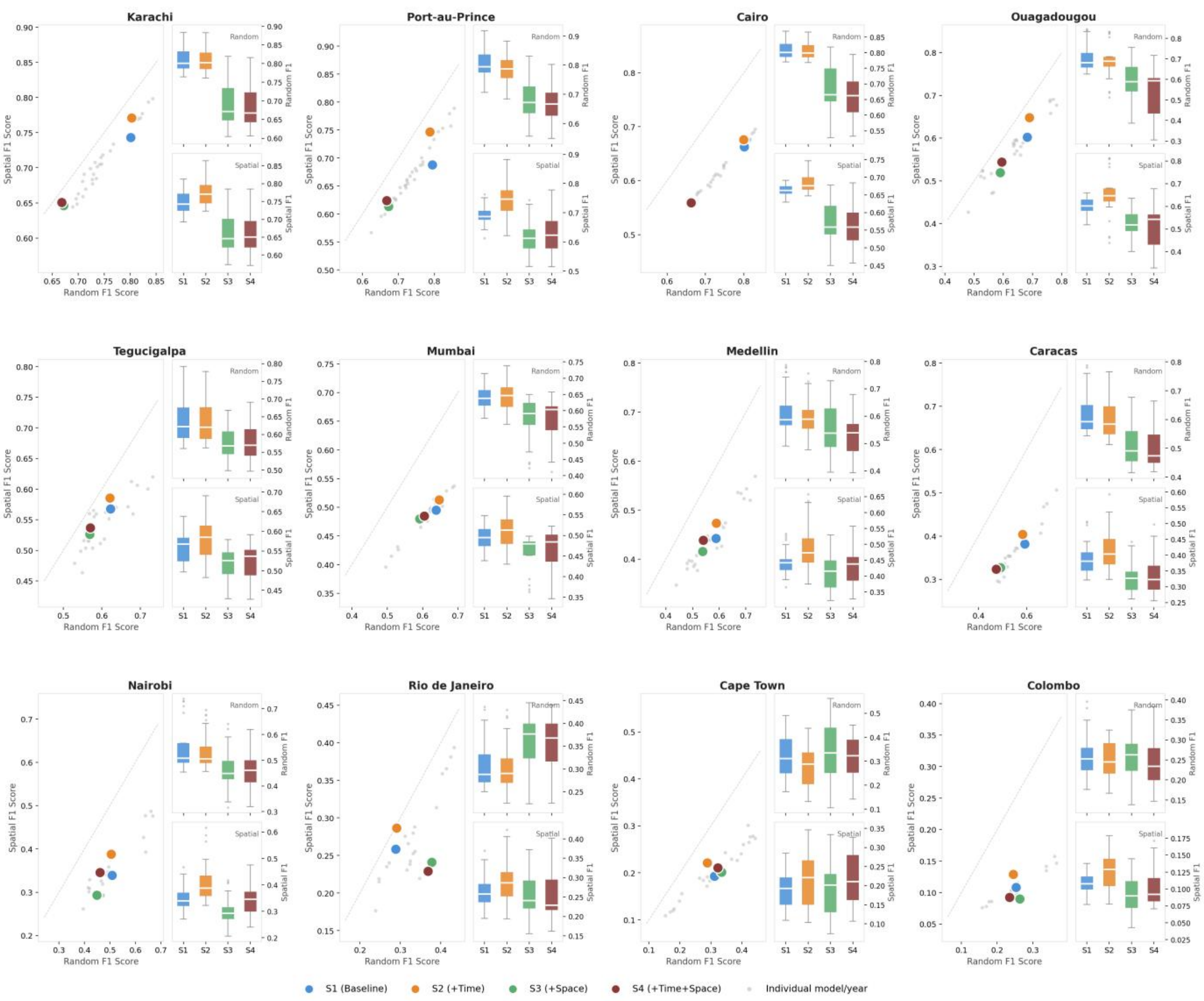

**Fig. 4 Per-city classification performance under S1–S4. Panels are ordered by descending S1 spatial F1 of TorchMLP.** Each panel: scatter of Random vs Spatial F1 across model–year combinations (gray dots) with strategy medians (colored circles) and the y = x diagonal; right column stacks Random F1 (top) and Spatial F1 (bottom) box plots across classifiers and years.

Among the strategies, S2 is the only one to outperform S1 comprehensively under spatial evaluation: the median spatial F1 of TorchMLP increases from 0.538 to 0.616 (+0.078), while HistGBT, RF, and LogReg improve by +0.046, +0.052, and +0.038, respectively. Cross-year training within the same city introduces temporal heterogeneity, serving as a form of time-based regularization, reducing the random→spatial gap for TorchMLP—the model with the largest drop—from 0.225 to 0.103. In contrast, S3 exhibits a lower median spatial F1 (0.500) than S1, indicating systematic distribution shifts when transferring AEF embeddings across cities. S4 partially recovers to 0.523 but remains below S2, demonstrating that merely increasing sample size cannot compensate for cross-city feature drift. This pattern, where temporal expansion is more beneficial than spatial expansion, aligns with findings in cross-city slum detection literature(Stark et al., 2020; Wurm et al., 2019).

At the city level (Figure 4), the median spatial F1 for S2 across the twelve cities spans 0.65 (Karachi 0.759 → Colombo 0.113), reflecting heterogeneity driven by the interaction of class proportion and spatial morphology.

- **Class proportion dominates extremes:** High-proportion cities (Karachi 22.6%, Port-au-Prince 22.3%, Cairo 20.1%) achieve stable spatial F1 > 0.65, whereas low-proportion cities (Cape Town 1.5%, Colombo 1.4%) drop to 0.21 and 0.11, with widened box spans approaching the lower stability limit under extreme class imbalance (positive-to-negative ratio ≈ 65:1; see Table A3 in the Appendix).
- **Spatial morphology dominates intermediate gradients:** Mumbai (8.3%, F1 = 0.498) performs worse than Tegucigalpa (11.6%, 0.568) and Ouagadougou (14.7%, 0.632), reflecting the difficulty of distinguishing multi-layer informal buildings adjacent to formal built-up areas at 10 m resolution. Rio de Janeiro (3.8%, 0.258) also underperforms relative to its class proportion, corresponding to scattered hillside favelas causing training-test distribution mismatch.

The improvement from S1→S2 also varies by city: Nairobi (+0.133) and Port-au-Prince (+0.101) benefit most from temporal regularization, while Mumbai (+0.022) and Rio de Janeiro (+0.029) show limited gains, indicating that temporal expansion cannot fully compensate for the intrinsic discriminative difficulty posed by spatial morphology. S3 and S4 fail to surpass S2 under spatial evaluation in all twelve cities.

### 4.1.2. Regression Task Performance

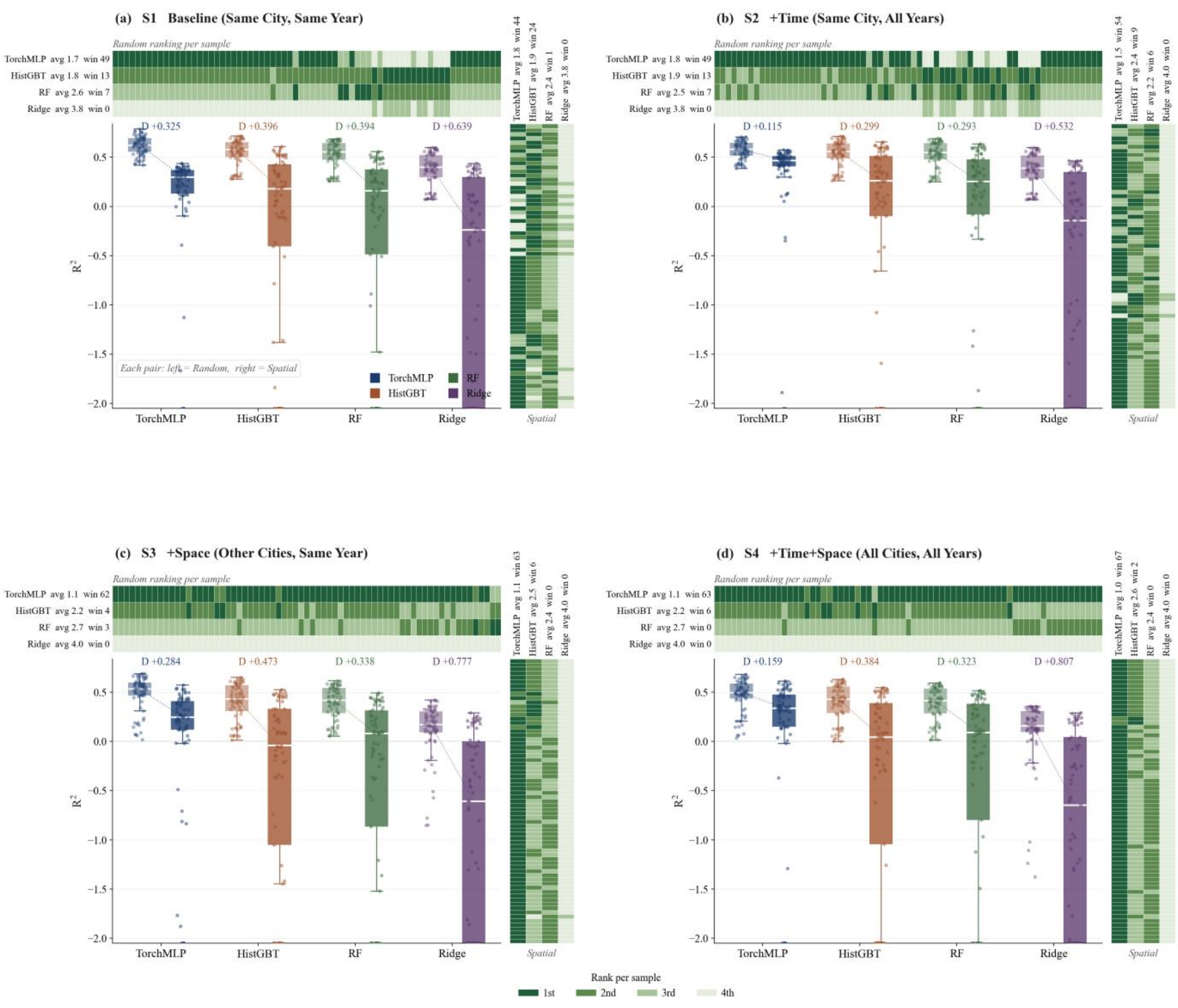

**Fig. 5 Regression $R^2$ across four training strategies and four models, evaluated on 69 city-year pairs.** (a) S1, baseline; (b) S2, +Time; (c) S3, +Space; (d) S4, +Time+Space. Within each model, paired box plots compare Random and Spatial evaluation, with the median Δ annotated. Top and right strips show per-sample model rankings under Random and Spatial evaluation; average rank and win count are labelled.

The relative performance pattern of the regression task across the four strategies is broadly consistent with that of the classification task, as shown in Figure 5: TorchMLP consistently ranks first, Ridge remains the weakest model, S2 comprehensively outperforms S1 under spatial evaluation, and S3 and S4 further deteriorate under cross-city transfer. However, regression is more sensitive to training–test spatial distribution mismatch than binary classification, leading to a substantially amplified random-to-spatial performance gap. Under S1, the Δ of TorchMLP increases from 0.225 in classification to 0.325 in regression; for HistGBT and RF, the gap increases from 0.11–0.13 to approximately 0.39; and for Ridge, it reaches 0.639. Ridge yields systematically negative $R^2$ values under spatial evaluation $(S1: -0.239; S3: -0.606; S4: -0.647)$, falling below the mean baseline. This contrasts with the classification task, where LogReg performs poorly but remains relatively stable, indicating spatial non-stationarity in the mapping between AEF embeddings and continuous density labels. The S1 spatial $R^2$values of HistGBT and RF are only 0.16–0.18, much lower than their classification spatial F1 scores of approximately 0.45, suggesting that regression places higher demands on model expressiveness.

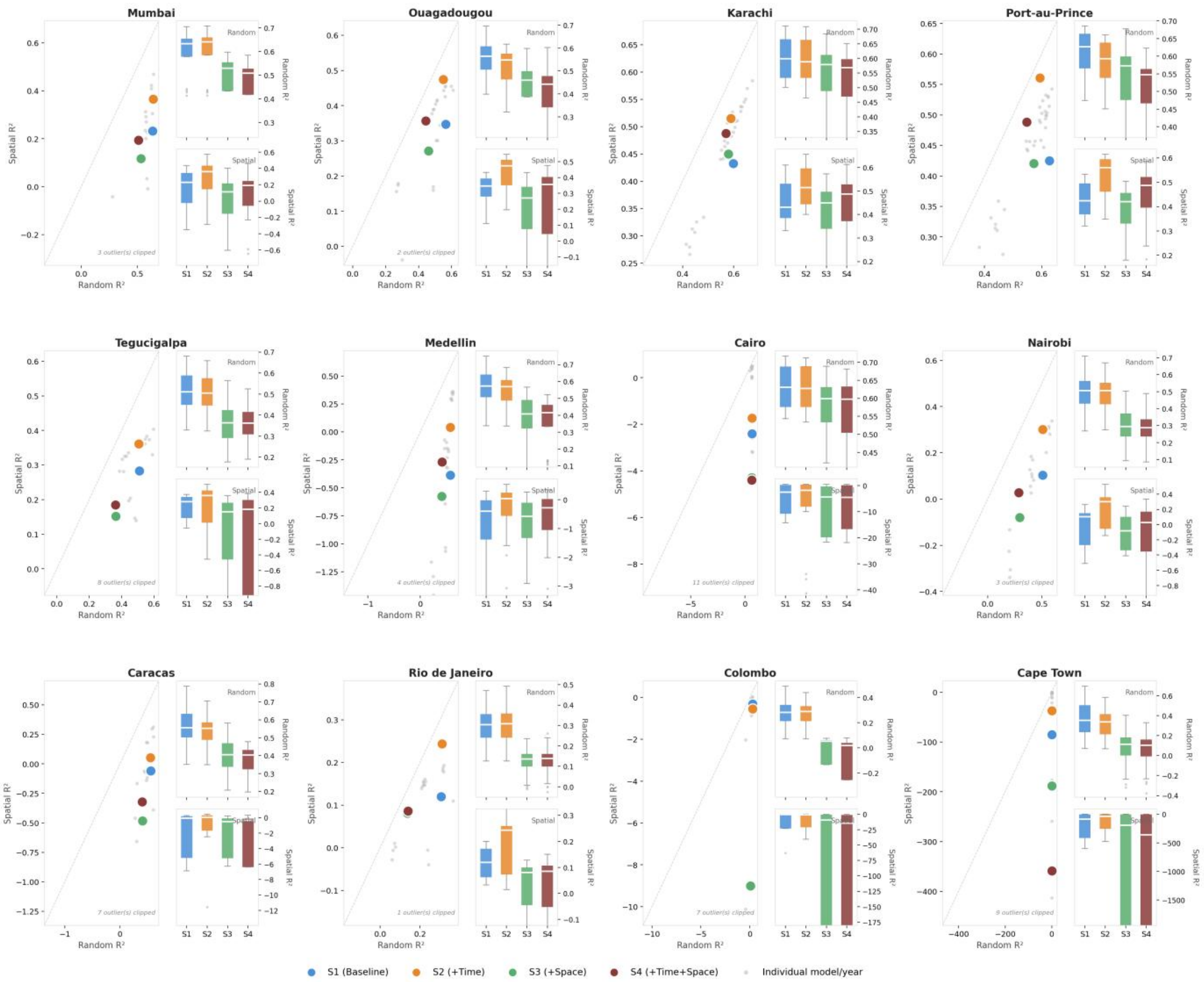

**Fig. 6 Per-city regression performance under S1−S4, ordered by descending TorchMLP S1 spatial $R^2$.** Within each panel, the scatter compares Random and Spatial $R^2$ across model‑year combinations (grey dots), with strategy medians shown as coloured circles and the y = x diagonal. The right column stacks Random (top) and Spatial (bottom) $R^2$ box plots across models and years. Axis ranges are set by IQR-based clipping; the number of clipped outliers is annotated.

The gains of S2 are larger for regression than for classification. For TorchMLP, the median spatial $R^2$increases from 0.299 under S1 to 0.466 under S2, corresponding to a gain of +0.167, compared with +0.078 in the classification task. Meanwhile, the random-to-spatial gap is reduced from 0.325 to 0.115. Under S3, the $R^2$of HistGBT becomes negative $(-0.041)$, while under S4, Ridge decreases to $-0.647$, indicating that cross-city negative transfer has a stronger effect on continuous-value prediction than on classification. The model ranking structure is similar to that observed in classification, although HistGBT and RF alternate more frequently between second and third place, with S1 spatial average ranks of 1.9 and 2.4, respectively.

At the city level, as shown in Figure 6, the median S1 spatial $R^2$of TorchMLP across the twelve cities spans approximately 3.4, which is much larger than the classification range of approximately 0.65. Moreover, the city rankings differ between the two tasks. Karachi, which ranks among the top cities in classification $(F1{=}0.759)$, ranks third in regression $(R^2{=}0.390)$, whereas Mumbai, which ranks sixth in classification ($F1$=0.498), achieves the highest regression performance ($R^2$=0.407). Both cities are characterized by high-density slum areas; however, Mumbai exhibits steep density gradients along slum boundaries, which can be captured by AEF embeddings. In contrast, Karachi contains relatively homogeneous corridor-like high-density areas, leaving limited exploitable density gradients. This ranking shift suggests that classification primarily emphasizes spatial boundary recognition, whereas regression depends more strongly on the continuity of intra-pixel density variation.

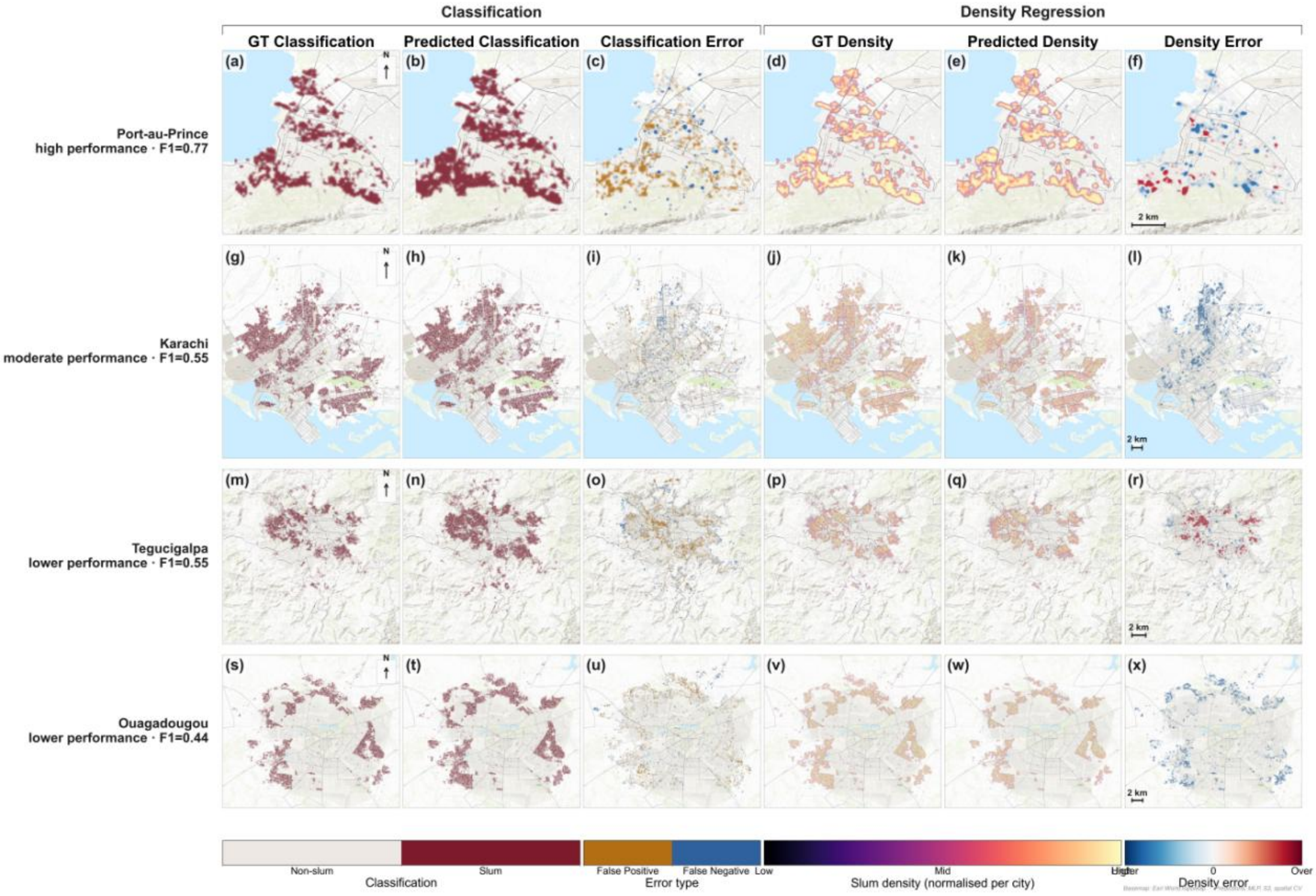

**Fig. 7 Spatial prediction patterns of slum classification and density regression for four cities under S2 (TorchMLP, AEF 64-dim, spatial CV), ordered by decreasing spatial F1: Port-au-Prince (0.77), Karachi (0.55), Tegucigalpa (0.55), Ouagadougou (0.44).** Columns: GT classification (a, g, m, s); predicted classification (b, h, n, t); classification error (c, i, o, u; orange = false positive, blue = false negative); GT density (d, j, p, v); predicted density (e, k, q, w); density error (f, l, r, x; red = over-prediction, blue = under-prediction). Overlays normalised per city. Basemap: Esri WorldTopoMap.

$R^2$ becomes numerically unstable in cities with very few positive samples. In Cape Town, where slum pixels account for only 1.5%, the median S1 spatial $R^2$ of TorchMLP is $-2.977$, with annual values fluctuating from $-68.99$ to $+0.04$; Ridge even decreases to $-287.53$ in this city. Colombo, with a similarly low slum proportion of 1.4%, shows a milder failure pattern ($R^2 =- 0.073$). Under extremely low positive-sample conditions, the denominator of $R^2$, namely the total variance, approaches zero, resulting in distorted metric scaling. This behavior does not occur in classification, where F1 still degrades stably within the range of 0.10–0.20 under extreme class imbalance. The transition from $S1 \rightarrow S2$ brings substantial recovery in regression performance: Cape Town improves by +2.846, from $-2.977$ to $-0.132$, while Nairobi, Rio de Janeiro, Caracas, and Medellín improve by +0.21–0.27.

### 4.1.3. Qualitative Analysis of Spatial Prediction Patterns

Figure 7 compares the classification and density maps of S2-TorchMLP predictions and GT labels for four representative cities, ordered by spatial F1 from high to low: Port-au-Prince, Karachi, Tegucigalpa, and Ouagadougou.

Classification predictions generally show a tendency toward patch expansion. Predicted boundaries often extend outward, adjacent patches are prone to merging, and false positives are concentrated along patch edges (Figure 7c, o, u), with this pattern being most pronounced in Ouagadougou (Figure 7u). Karachi represents a contrasting case, where errors are dominated by blue false negatives that are spatially concentrated along several linear belt-like areas (Figure 7i). This suggests that AEF embeddings systematically under-detect linear slum settlements extending along transportation corridors.

The spatial pattern of density errors is highly coupled with classification errors. False-negative areas correspond to density underestimation (Figure 7l, r, x), whereas false-positive areas correspond to density overestimation, particularly in the central area of Figure 7r. These results indicate that classification-stage biases are propagated to density estimation through the two-stage framework. Port-au-Prince exhibits the most spatially dispersed and lowest-magnitude density errors (Figure 7f), consistent with its higher spatial F1. In contrast, density errors in the other three cities are concentrated in specific spatial blocks. Overall, AEF embeddings localize compact high-density slum settlements relatively accurately, but their ability to identify linear corridor-type and low-density dispersed morphologies remains limited.

#### 4.1.4. Decomposition of Regression $R^2$

The S2 spatial $R^2$ reported in section 4.1.2 reflects the overall predictive accuracy of AEF embeddings for slum density regression. However, given the strongly bimodal distribution of density labels, with 89.3% of pixels being zero as shown in section 2.2.2, the diagnostic meaning of this metric requires further decomposition. This section uses the three diagnostic metrics defined in section 3.3.2, namely two-stage gain, oracle gain, and positive-pixel $R^2(\text{pos } R^2)$, to identify the actual sources driving $R^2$performance (Table 5).

Across the twelve cities, the median two-stage gain is 0, with a range from $-0.001$ to $+0.001$, indicating that the hard-cascaded two-stage framework provides no observable improvement over single-stage regression. This is because the single-stage TorchMLP already produces near-zero predictions for the 89.3% of zero-density pixels, leaving little marginal room for explicit zeroing. Replacing the classifier predictions with the true $y_{\text{cls}}$ for oracle zeroing yields only a median gain of $+0.038$, with an IQR of 0.018–0.058 and a city-level range from $+0.013$to $+0.105$. Even with an ideal classifier, $R^2$ can increase by only approximately 0.04. This result rules out the optimization pathway in which improving classification performance alone would substantially improve density estimation. In high-classification-performance cases, where classification F1 ranges from 0.69 to 0.84 (EGY, HTI, PAK, IND), the oracle gain is mostly below 0.03. In contrast, cases with lower classification F1 (ZAF, BRA, LKA) show relatively higher oracle gains of 0.06–0.07, but their single-stage $R^2$values are already low, meaning that even ideal classification would not raise density regression to a moderate performance level.

The diagnostic results for the positive-pixel subset are more revealing. The pos $R^2$is consistently negative across all twelve cities, ranging from $-0.28$ to $-4.7$, with a median of $-2.00$. This indicates that, for non-zero pixels, TorchMLP using AEF embeddings alone fails to outperform the mean baseline in density prediction. Together, the three diagnostic metrics show that the $R^2$reported in section 4.1.2 is primarily driven by the ability of AEF embeddings to distinguish zero from non-zero slum pixels, consistent with the spatial boundary clarity observed in the classification task in section 4.1.1. By contrast, the capacity to model continuous density variation within positive pixels remains limited under the current representation.

**Table 5. Per-city diagnostic decomposition of regression $R^2$ under S2 with TorchMLP (spatial CV), sorted by descending single $R^2$.** cls F1 and single $R^2$ are reported as per-city cross-year medians; the bottom row reports cross-city medians for the three diagnostic quantities only, as per-city descriptors are not meaningful to aggregate. n_yr, number of GT-labelled years per city.

| City | n_yr | cls F1 | single R² | Two-stage gain | Oracle gain | pos R² |
|---|---|---|---|---|---|---|

| City | n_yr | cls F1 | single $R^2$ | Two-stage gain | Oracle gain | pos $R^2$ |
|---|---|---|---|---|---|---|
| IND | 4 | 0.69 | 0.67 | 0 | 0.029 | -1.26 |
| HON | 6 | 0.72 | 0.6 | 0 | 0.038 | -0.28 |
| HTI | 8 | 0.84 | 0.58 | 0 | 0.017 | -2.19 |
| BFA | 6 | 0.8 | 0.57 | -0.001 | 0.013 | -3.81 |
| EGY | 6 | 0.84 | 0.55 | 0 | 0.018 | -4.68 |
| KEN | 5 | 0.61 | 0.54 | 0 | 0.054 | -1.04 |
| COL | 7 | 0.64 | 0.53 | 0 | 0.045 | -1.39 |
| PAK | 7 | 0.83 | 0.53 | 0 | 0.014 | -3.76 |
| VEN | 5 | 0.57 | 0.44 | 0 | 0.105 | -2.01 |
| ZAF | 6 | 0.34 | 0.35 | 0 | 0.058 | -3.82 |
| BRA | 5 | 0.35 | 0.27 | 0 | 0.066 | -3.15 |
| LKA | 4 | 0.15 | 0 | 0 | 0.059 | -1.69 |
| **Median** | — | — | — | **0** | **0.038** | **−2.00** |

This distinction has important implications for practical applications. For tasks focused on slum presence or absence, or on coarse coverage-level stratification, such as SDG 11.1 monitoring and preliminary screening for urban spatial planning, the 64-dimensional AEF embeddings combined with the standard models tested in this study can already provide useful classification signals. However, for tasks requiring accurate characterization of intra-pixel density gradients, such as quantifying infrastructure service radii or supporting fine-grained density-based planning, AEF embeddings alone are insufficient. Additional information sources may be required, including nighttime lights, remote sensing indices, spatial structure, and POI-based auxiliary factors, whose integration effects are systematically evaluated in section 4.3. Alternatively, more density-sensitive modeling paradigms may be needed, such as adopting a Mixture-of-Experts design similar to GRAM to disentangle city-specific signals, or incorporating higher-resolution local imagery to support density-gradient modeling.

## 4.2. Dimensional Feature Analysis

### 4.2.1. Principal Component Analysis

PCA ablation was applied to the 64-dimensional AEF embeddings, with $k \in \{8,16,24,32,38,48,56,64\}$. The classification and regression tasks exhibit systematic differences in their dimensional requirements (Figure 8).

For the classification task, performance reaches statistical saturation at the aggregate level when $k = 32$, corresponding to a cumulative explained variance of 98.1%. At this dimensionality, the median spatial F1 scores of all four models reach 98–101% of their full-dimensional performance ($k$=64). In the Wilcoxon paired tests, HistGBT ($p$=0.263)and RF ($p$=0.238)show no significant difference between $k = 32$ and $k = 64$. Although the differences remain statistically significant for TorchMLP and LogReg ($p$<0.001), the effect sizes are small, with $\Delta =+ 0.018$ and $\Delta =+ 0.015$, respectively. For HistGBT and RF, $\Delta$ becomes negative in the range of $k = 38 – 56$, suggesting that low-variance trailing principal components may introduce slight noise into tree-based models.

At the city level (Figure 9a,b), the individual saturation point $k^*$ for most cities falls within $k = 24 – 32$. Cairo reaches saturation as early as $k = 16$, whereas Cape Town and Colombo require $k \geq 48$. The conclusions are consistent between random and spatial evaluation, indicating that $k = 32$is a relatively conservative aggregate threshold for classification.

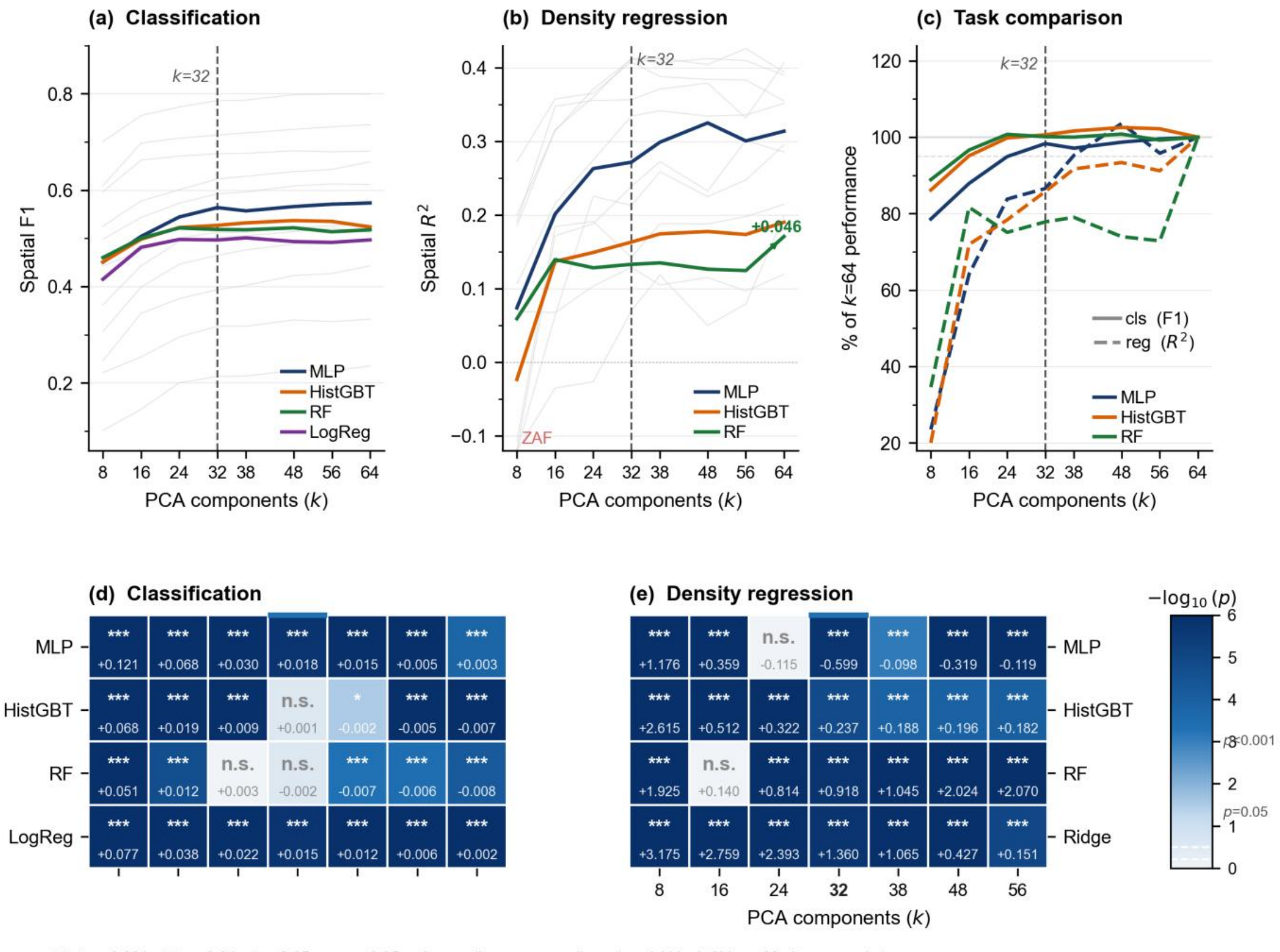

**Fig. 8 PCA component ablation under S1 (spatial block CV, n = 69 city−year pairs).** (a) Spatial F1 and (b) spatial $R^2$ as a function of retained PCA components k, for four classifiers and three regressors (per-city MLP curves in grey; ZAF curve = Cape Town, AEF near-failure case). (c) Performance as percentage of the full k = 64 baseline. (d, e) Wilcoxon signed-rank test comparing each k against k = 64 across four classifiers / four regressors (Ridge included in e only); colour encodes $-\log_{10}(p)$, cell values give the median Δ. Dashed line at k = 32 marks 98.1% cumulative variance explained.

In contrast, density regression remains unsaturated even at $k = 64$. Most Wilcoxon comparisons are highly significant (Figure 8e), except for TorchMLP at $k = 24$ and RF at $k = 16$, which are not significant. At $k = 32$, the spatial $R^2$ values of TorchMLP, HistGBT, and RF reach only 86.6%, 85.8%, and 77.8% of their full-dimensional performance, respectively. RF stagnates over the range of $k = 24 - 56$, followed by a sharp increase of $\Delta = +0.046$ from $k = 56$ to $k = 64$, suggesting that the highest-order principal components still carry additional information for density estimation. The $R^2$ curve of TorchMLP is non-monotonic, with a local peak at $k = 48$, reflecting stochastic fluctuations in neural network optimization.

At the city level (Figure 9c,d), under spatial cross-validation, $k^*$ is generally delayed to $k = 32$–$64$. Cape Town and Colombo show no valid saturation point because their $R^2$ values remain $\leq 0$ throughout the entire dimensional range. This is consistent with the aggregate analysis, where these two cities are identified as failure cases for AEF-based density regression.

### 4.2.2. SHAP Analysis

SHAP analysis further identifies the specific principal components that support the dimensional differences described above (Figure 10a,b). In the classification task, PC36, PC8, and PC34 are consistently selected into the top 10 by all four classifiers, showing full consensus $(4/4)$. PC36 ranks first in TorchMLP and LogReg, whereas PC8 ranks first in HistGBT and RF.

In the regression task, PC36 and PC16 both achieve full consensus $(4/4)$. Notably, PC36 ranks first across all four regression models. Across both tasks, PC36 is the most stable dominant dimension in the AEF 64-dimensional embedding: it ranks first in all four regression models and in two classification models, namely TorchMLP and LogReg, and ranks second in the HistGBT and RF classification models.

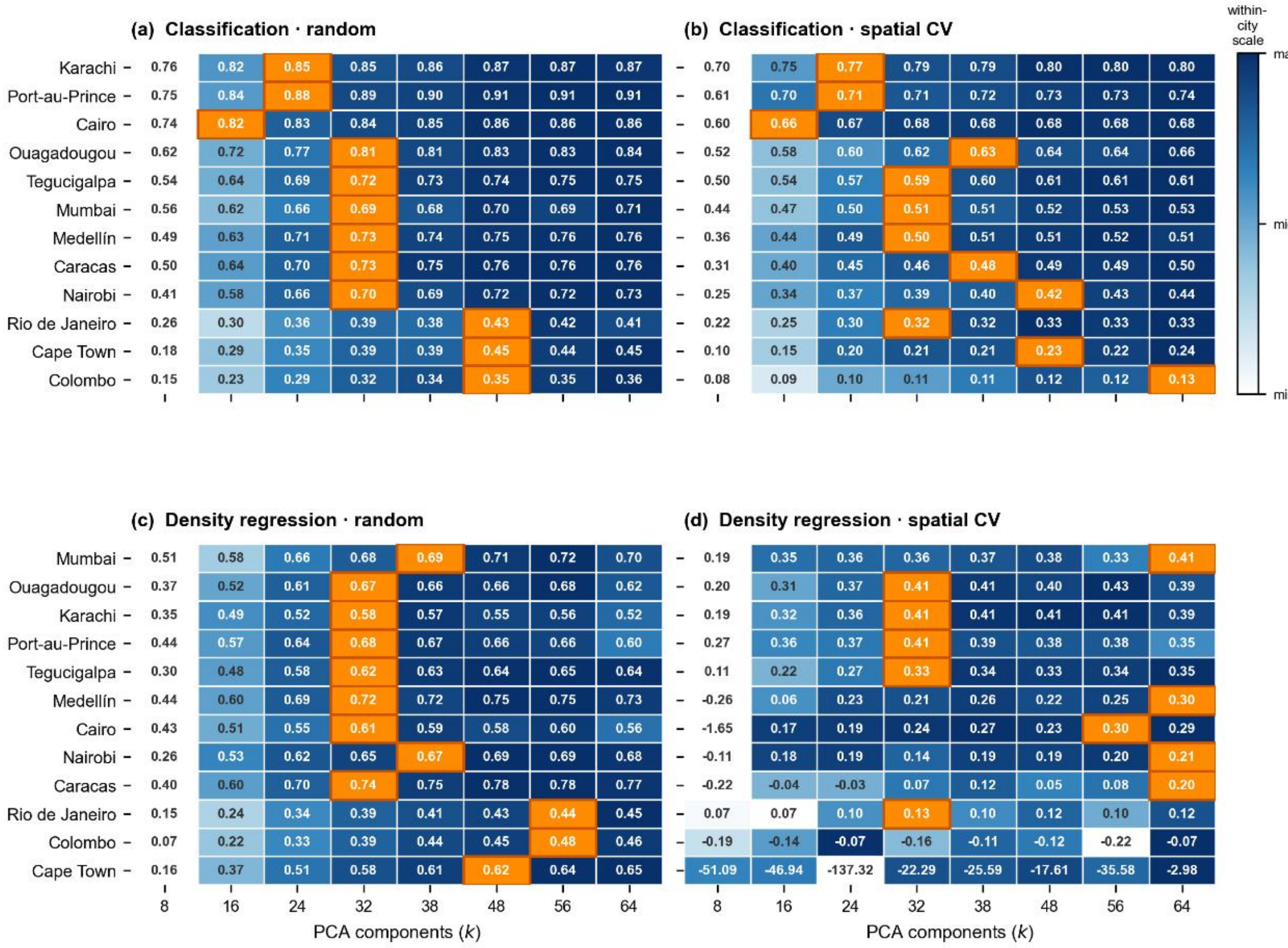

**Fig. 9 City-level PCA saturation profiles for MLP under S1, ordered by descending performance at k = 64.** (a, b) Absolute spatial F1 for classification under random and spatial CV; (c, d) absolute spatial $R^2$ for density regression under random and spatial CV. Colour encodes each city's row-normalised scale (light = min, dark = max). Orange cells mark k*, the smallest k reaching ⩾ 95% of that city's maximum; cities with maximum $R^2 \leqslant 0$ (Cape Town, Colombo in spatial CV) are not highlighted.

The secondary consensus dimensions show task-specific differentiation. Classification relies more strongly on PC8 and PC34, whereas regression emphasizes PC16 and PC26, with no overlap between the two groups. This division of labor corresponds to the pattern observed in section 4.2.1: classification saturates early because a small number of highly consistent dimensions provide sufficient discriminative information, whereas density estimation requires a broader subset of principal components.

At the city level (Figure 10c), PC36 ranks first in the classification task for most cities, including Karachi, Cairo, Ouagadougou, Caracas, and Rio de Janeiro. However, in Port-au-Prince and Mumbai, dimensions such as PC8, PC16, and PC1 become relatively more prominent, reflecting city-specific dependence on different principal components associated with slum morphology. In the regression panels, Cape Town and Colombo show markedly fewer prominent SHAP bubbles, consistent with the failure cases reported in section 4.1.2, where both cities have spatial $R^2 \leq 0$. This indicates that the AEF principal components do not effectively encode density-gradient signals in these two cities.

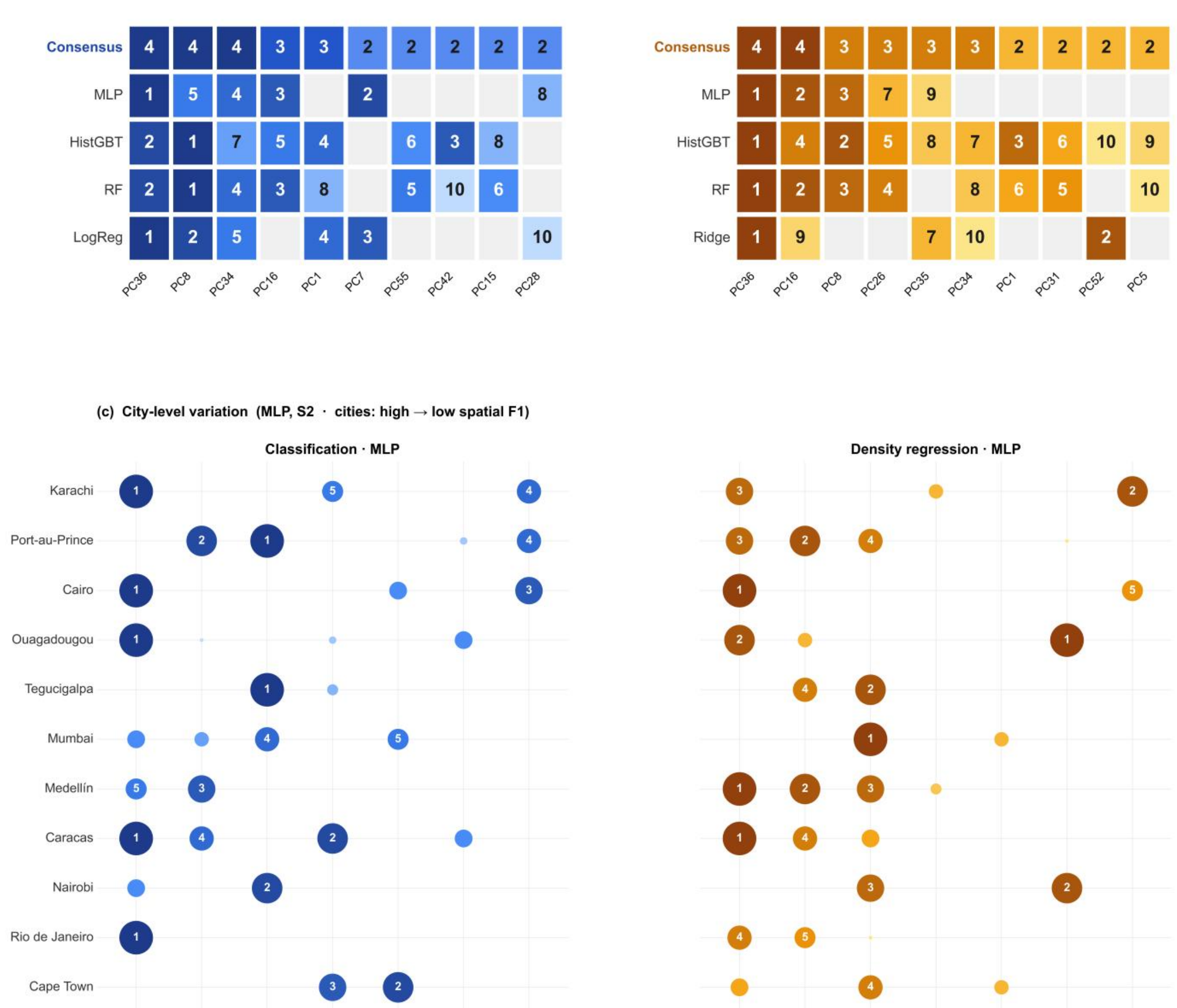

**Fig. 10 SHAP importance of AEF principal components under S2, top 10 dimensions per model.** (a, b) Cross-model consensus grids for classification and density regression; numbers give each model's importance rank for that PC (1 = highest), colour depth encodes rank; the Consensus row counts models selecting each PC. Columns ordered by descending consensus then ascending mean rank. (c) Per-city importance ranks (MLP only) for the eight PCs with consensus $\geqslant$ 3; bubble size reflects rank (larger = higher); only rank $\leqslant$ 5 labelled; cities ordered by descending k = 64 spatial F1; grey cells = PC outside that city's top 10.

### 4.3. Effects of Auxiliary Factor Integration

#### 4.3.1. Overall Effects

Figure 11 presents the distributions of $\Delta F1$, $\Delta IoU$, and $\Delta R^2$ for the five auxiliary factor combinations ($C1$–$C5$)relative to the C0 baseline, with random and spatial evaluations shown side by side.

C1 ($AEF$+NTL)contributes almost no systematic improvement beyond the AEF embeddings, with spatial $\Delta F1 =+ 0.001$, $\Delta IoU =+ 0.002$, and $\Delta R^2 =+ 0.002$. Within cities, slums are often among the weakest nighttime-light areas because of limited access to formal electricity grids, reliance on temporary lighting, and the mismatch between high population density and low light output. As a result, NTL features struggle to distinguish slums from other low-light areas, such as vacant land, buffer zones, and peri-urban villages, and therefore do not provide a strongly discriminative signal for slum detection. A few negative outliers in the spatial subset, where some cities show $\Delta R^2 <- 0.10$, may arise because NTL captures other intra-urban semantics, such as commercial versus residential activity, which can introduce additional confusion.

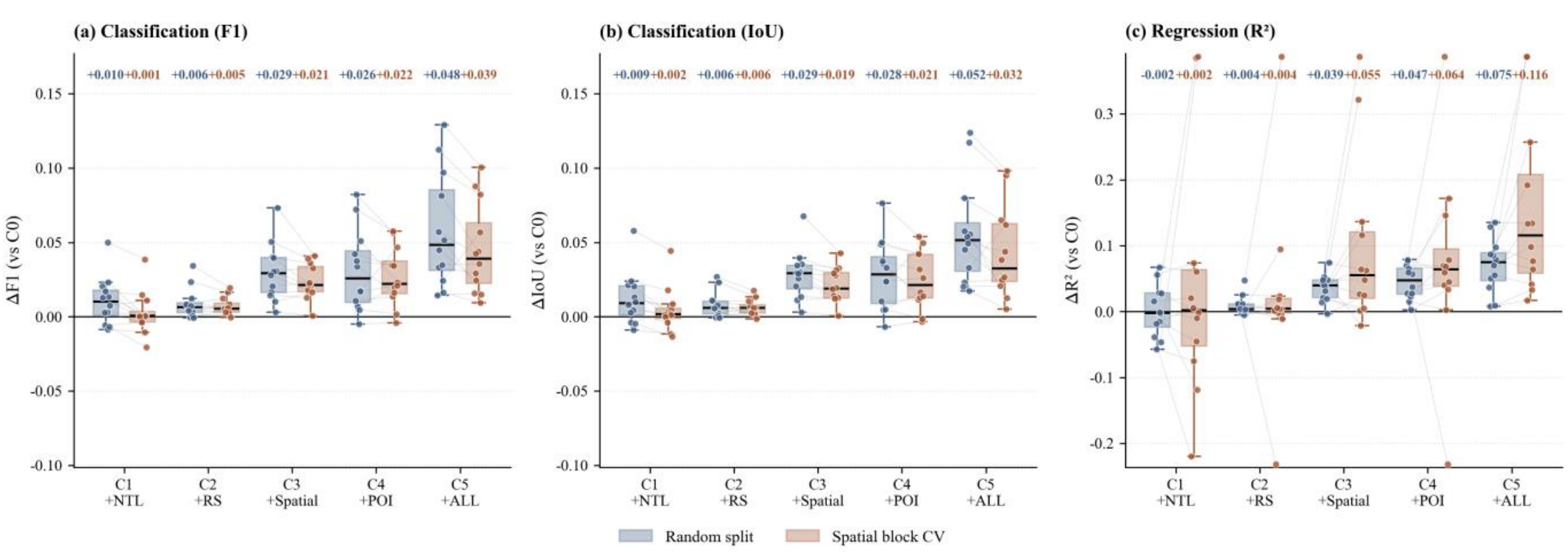

**Fig. 11 Performance gains of five auxiliary feature configurations (C1−C5) over the C0 (pure AEF) baseline across 12 cities (n = 69 city−year pairs).** (a) ΔF1, (b) ΔIoU, (c) $\Delta R^2$. Each dot is one city; grey lines pair its Random and Spatial values. Box plots show inter-quartile range and median; numbers above each box give the median Δ for Random (blue) and Spatial CV (orange). Y-axis ranges set by IQR-based bounds; out-of-range values clipped to the panel edge.

C2 ($AEF$+RS) produces a small but stable positive gain, with spatial $\Delta F1 =+ 0.005$, $\Delta IoU =+ 0.006$, and $\Delta R^2 =+ 0.004$. It also shows the narrowest box range across the twelve cities, indicating the highest cross-city stability. The combination of "low vegetation and high built-up intensity" captured by NDVI, NDBI, and GHSL represents a typical physical signature of slum areas. However, its contribution is constrained by a fundamental ceiling: formal high-density built-up areas, including compact old urban districts and dense residential cores, often exhibit similar spectral and built-up characteristics. Thus, RS factors can identify whether an area is densely built, but they cannot independently distinguish formal from informal built-up environments.

C3 ($AEF$+Spatial) and C4 ($AEF$+POI) show comparable contributions to classification metrics, with spatial $\Delta F1 \approx+ 0.021$ and $\Delta IoU \approx+ 0.019$. Their effects are substantially stronger for regression, with $\Delta R^2 =+ 0.055$ for C3 and $\Delta R^2 =+ 0.064$ for C4. Their contribution to density estimation is approximately 2.6–3.0 times their contribution to boundary recognition. This ratio reflects a task-level semantic distinction: AEF can already identify slum boundaries from visual texture, whereas modeling density gradients requires structured evidence related to location and service provision. Spatial factors, such as distance to major roads, slope, and elevation, describe typical locational patterns of slums, including settlements on urban fringes, steep slopes, flood-prone areas, railway corridors, and areas near landfills. POI factors encode relative deprivation in formal social services, as the densities of banks, post offices, formal hospitals, and chain businesses are generally lower in slums than in surrounding areas. The standalone gain of C4 in regression, $\Delta R^2 =+ 0.064$, is the largest among individual auxiliary factors, consistent with the systematic scarcity of formal POIs in slum areas. This type of semantic information is difficult for AEF visual embeddings to infer directly from texture alone.

C5 ($AEF$+ALL) achieves the largest median improvement across all three metrics, with spatial $\Delta F1 =+ 0.039$, $\Delta IoU =+ 0.032$, and $\Delta R^2 =+ 0.116$. This indicates additive synergy among the auxiliary factors: the $\Delta R^2$ of C5 is approximately 1.8 times that of C4 alone. RS, Spatial, and POI factors respectively encode the physical appearance, locational structure, and social-service deprivation of slums, and their complementarity helps construct a more complete feature profile. The slight interference introduced by NTL is offset by the other factors. However, C5 also exhibits the widest distribution, with spatial $\Delta F1$ ranging from $+0.005$ to approximately $+0.10$, indicating pronounced inter-city heterogeneity. Cities whose slum characteristics align across the three dimensions of physical form, location, and social-service deprivation benefit most from full-factor integration. In contrast, for cities with atypical morphologies, the directions of the three signals may diverge, weakening their synergistic effect.

In addition, for the regression task, the gains under spatial evaluation are generally larger than those under random evaluation for C3, C4, and C5, which is opposite to the pattern observed for classification metrics. This suggests that auxiliary factors provide especially strong support for density estimation under spatial extrapolation. When AEF visual embeddings degrade under cross-city transfer, RS-based physical indicators, Spatial locational regularities, and POI-based service-deprivation features act as more transferable low-level semantic signals, thereby providing stable compensation for density prediction.

### 4.3.2. City-Level Heterogeneity

The classification and regression tasks exhibit systematic divergence in their optimal auxiliary configurations (Figure 12). For the classification metrics ($\Delta F1/\Delta IoU$), C5 performs best in 11 of the 12 cities. The magnitude of improvement is inversely related to the baseline slum proportion: BFA, KEN, and ZAF achieve $\Delta F1$values of approximately $+0.08–+0.10$, whereas PAK and HTI are constrained by ceiling effects, with gains of only $+0.036–+0.057$. EGY shows the smallest improvement ($\leq + 0.02$). The only exception is HON (Tegucigalpa), where C3 performs best, consistent with the dominant role of locational features in classification discrimination under mountainous slope conditions.

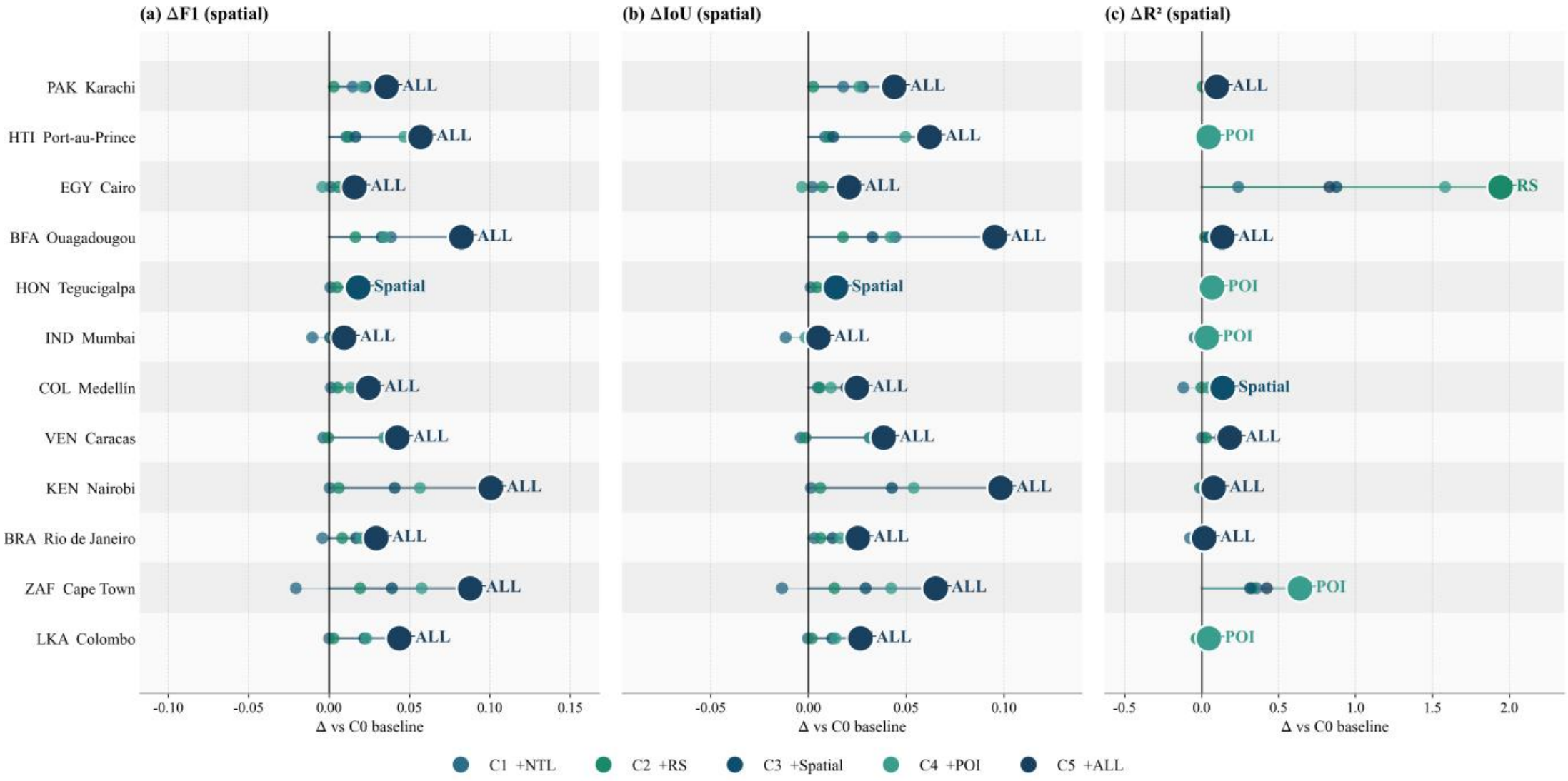

**Fig. 12 Per-city auxiliary feature gains over the C0 (pure AEF) baseline under S2 (TorchMLP, spatial block CV). (a) ΔF1, (b) ΔIoU, (c) ΔR².** Rows: 12 cities ordered by descending S2 spatial F1 of the C0 baseline. For each city and metric, dots show the median Δ of each configuration; the per-city best configuration is enlarged and labelled. Negative values indicate degradation. Note the wider x-axis in (c), reflecting greater city-level heterogeneity in regression.

The optimal configuration pattern for the regression task ($\Delta R^2$)is markedly different. C4 performs best in five cities, namely HTI, HON, IND, ZAF, and LKA. C5 wins in another five cities, namely PAK, BFA, KEN, VEN, and BRA. C3 and C2 each perform best in one city, COL and EGY, respectively. EGY and ZAF are two prominent outliers with substantially larger gains: EGY achieves $\Delta R^2 \approx + 1.9$ under C2, while ZAF achieves $\Delta R^2 \approx + 0.6$ under C4. These patterns correspond respectively to the strong discriminative power of built-up density indices for Cairo's linear corridor-type slums and the sharp contrast between formal POI distributions and slum areas under Cape Town's segregated urban structure. C1consistently lies at the lower end across all cities and all metrics, with the strongest negative deviation observed in ZAF. The cross-city dominance of C5 in classification does not extend to density estimation, where single-factor configurations, particularly C4 or C3, often perform best. Therefore, auxiliary factor selection should be differentiated by task type.

Table 6 reports the absolute performance values under the optimal configurations. Across the twelve cities, the median spatial F1 increases from 0.476 under C0 to 0.493 under the optimal configurations, while the cross-city ranking remains largely unchanged. High-baseline cities obtain only modest gains of $+0.036$–$+0.057$, whereas mid- to low-baseline cities, such as KEN and ZAF, receive larger absolute compensation of $+0.088 - +0.100$. The recovery is more pronounced for the regression task. EGY ($\Delta R^2 =+ 1.943$) and ZAF ($\Delta R^2 =+ 0.638$) shift from negative to positive $R^2$ values, indicating targeted compensation for Cairo's corridor-type slums and Cape Town's segregated slum pattern.

Using $F1 \geq 0.5$ and $R^2 \geq 0.3$ as the minimum usability thresholds for the joint classification–density task, corresponding respectively to practical discrimination between positive and negative classes and density estimation clearly outperforming the mean baseline, six cities reach the dual-task usability threshold under their optimal auxiliary configurations: PAK, HTI, BFA, EGY, HON, and IND. KEN meets only the regression threshold $(F1 = 0.470 < 0.5)$. The remaining five cities, VEN, COL, BRA, ZAF, and LKA, fail to meet both thresholds. Among them, LKA $(F1 = 0.156, R^2 =- 0.002)$ remains close to a failure state even after full auxiliary factor integration. For such cities, the extremely low positive-class proportion (≤1.5%)and atypical slum morphology impose structural constraints that exceed the compensatory capacity of the current auxiliary factor system. These cases likely require higher-resolution local imagery or more density-sensitive modeling paradigms.

**Table 6. Per-city absolute performance under the best auxiliary configuration in classification (F1) and regression ($\mathbf{R^2}$) tasks (S2, TorchMLP, spatial block CV).** C0 columns report the AEF 64-dim baseline; *_best columns give the highest value across C1 – C5 with the corresponding configuration label. Cities ordered by descending F1_C0.

| Code | City | F1_C0 | F1_best_cfg | F1_best | dF1 | R2_C0 | R2_best_cfg | R2_best | dR2 |
|---|---|---|---|---|---|---|---|---|---|
| PAK | Karachi | 0.759 | +ALL | 0.794 | 0.036 | 0.478 | +ALL | 0.576 | 0.098 |
| HTI | Port-au-Prince | 0.716 | +ALL | 0.773 | 0.057 | 0.507 | +POI | 0.552 | 0.045 |
| BFA | Ouagadougou | 0.632 | +ALL | 0.715 | 0.082 | 0.431 | +ALL | 0.564 | 0.134 |
| EGY | Cairo | 0.671 | +ALL | 0.687 | 0.016 | -1.582 | +RS | 0.362 | 1.943 |
| HON | Tegucigalpa | 0.568 | +Spatial | 0.586 | 0.018 | 0.333 | +POI | 0.4 | 0.068 |
| IND | Mumbai | 0.498 | +ALL | 0.508 | 0.009 | 0.322 | +POI | 0.356 | 0.034 |
| KEN | Nairobi | 0.37 | +ALL | 0.47 | 0.1 | 0.269 | +ALL | 0.346 | 0.077 |
| VEN | Caracas | 0.387 | +ALL | 0.429 | 0.042 | 0.064 | +ALL | 0.246 | 0.182 |
| COL | Medellín | 0.453 | +ALL | 0.477 | 0.024 | -0.03 | +Spatial | 0.107 | 0.136 |
| BRA | Rio de Janeiro | 0.258 | +ALL | 0.287 | 0.029 | 0.201 | +ALL | 0.218 | 0.017 |
| ZAF | Cape Town | 0.208 | +ALL | 0.296 | 0.088 | -0.444 | +POI | 0.194 | 0.638 |
| LKA | Colombo | 0.113 | +ALL | 0.156 | 0.044 | -0.049 | +POI | -0.002 | 0.046 |

#### 4.3.3. Spatial Evidence of Auxiliary-Factor Repair Patterns

Figure 13 compares pixel-wise errors between C0 and the optimal auxiliary configurations using 2022 data from four cities: PAK, HTI, BFA, and HON. The optimal configurations are $+\mathrm{ALL}$ for PAK, HTI, and BFA, and $+\mathrm{Spatial}$ for HON. The third and sixth columns show the spatial improvement or degradation introduced by auxiliary factors, calculated as $|\mathrm{Error}_{C0}| - |\mathrm{Error}_{Best}|$, where blue indicates improvement and red indicates degradation.

The repair patterns are highly heterogeneous across the four cities. In Karachi (Figure 13a–f),

classification improvements are distributed across the city's secondary road network, while density improvements are concentrated along the port and major transportation corridors (Figure 13f). This pattern is consistent with the corridor-type morphology of slums in Karachi and the limited gain of $dR^2 =+ 0.098$ achieved by $+$ALLunder spatial cross-validation. In Port-au-Prince (Figure 13g–l), the density difference map shows concentrated red overestimation patches in the central urban area (Figure 13l), corresponding to the marginal gain of $dR^2 =+ 0.045$ obtained in this city. Although auxiliary factors improve overall $R^2$ , they also introduce localized overestimation in the high-density port area.

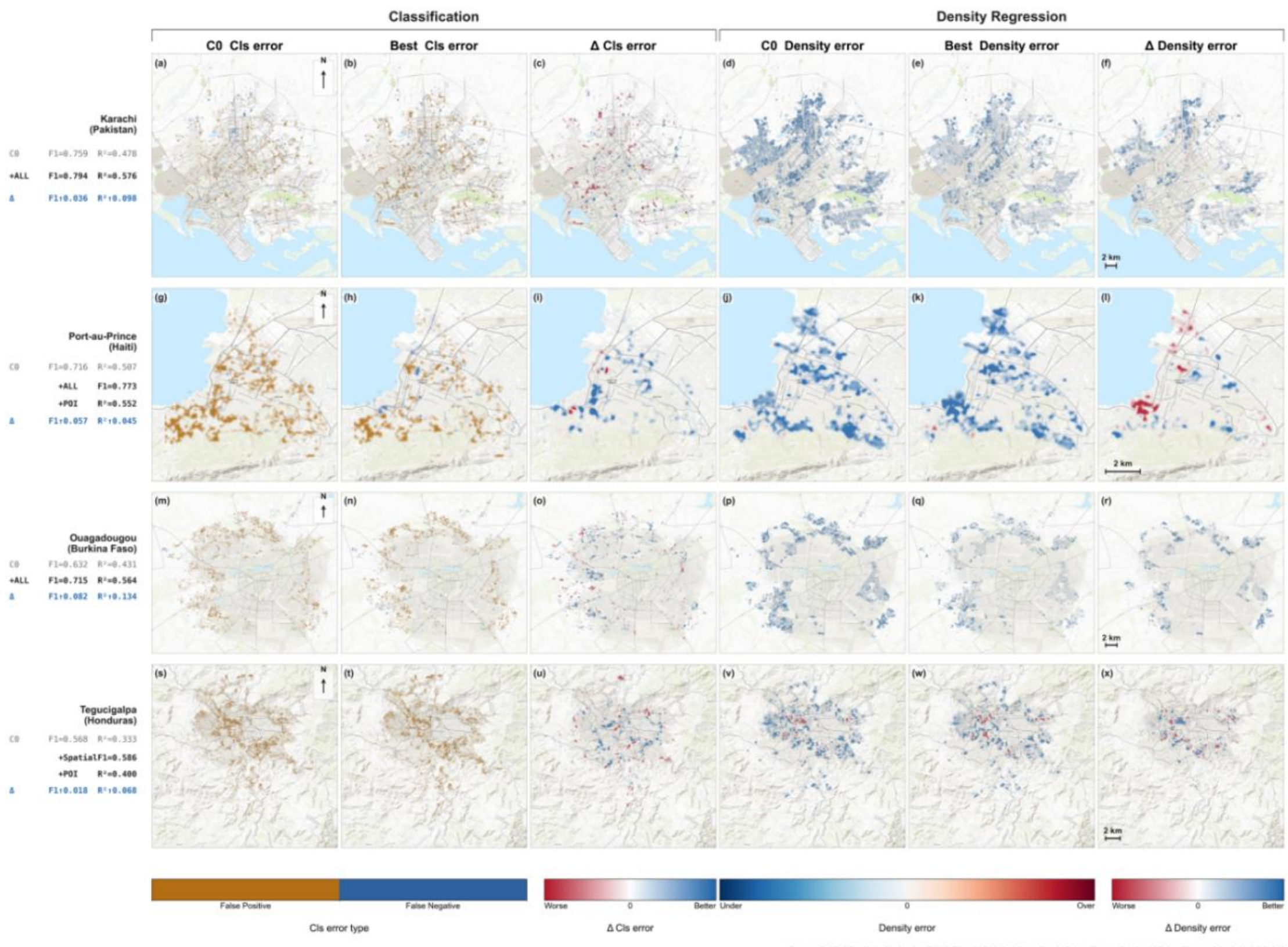

**Fig. 13 Pixel-wise spatial decomposition of auxiliary factor contribution at four representative cities (2022): Karachi, Port-au-Prince, Ouagadougou, Tegucigalpa, ordered by descending C0 spatial F1.** Columns: classification error type (a, g, m, s for C0; b, h, n, t for best configuration; orange = false positive, blue = false negative); Δ classification error (c, i, o, u; |C0| − |Best|, blue = improvement, red = degradation); density error (d, j, p, v for C0; e, k, q, w for best; blue = under-prediction, red = over-prediction); Δ density error (f, l, r, x). Δ panels thresholded at $|\Delta| > 0.15$ to highlight patterns above noise. Predictions: TorchMLP under S2 with $3 \times 3$ spatial block CV. Basemap: Esri WorldTopoMap.

In Ouagadougou (Figure 13m–r), all four difference panels show relatively uniform blue improvements, with $dF1 =+ 0.082$ and $dR^2 =+ 0.134$, the largest gains among the four cities. This suggests that $+$ALL produces broad and stable compensation in Ouagadougou without obvious negative interactions. In Tegucigalpa (Figure 13s–x), density improvements (Figure 13v–x) are concentrated in the city center and eastern hillside settlement areas. This indicates that $+$Spatial compensates for the limitations of AEF visual embeddings in mountainous slope-type slums by encoding locational deprivation gradients, consistent with the city-level finding in section 4.3.2 that HON is the only city where $+$Spatial, rather than $+$ALL, performs best.

The repair effects of auxiliary factors are therefore not spatially uniform. Instead, they provide localized compensation for specific spatial morphologies, including transportation corridors in PAK, urban fringe expansion zones in BFA, and hillside locational deprivation in HON. This spatial heterogeneity supports the conclusion in section 4.3.2 that auxiliary-factor selection should be adapted to city-specific slum morphology.

## 5. Application Extension and Spatial Structure Validation

Section 4 validated the discriminative capacity of AEF for the dual slum-mapping tasks under spatial cross-validation from three perspectives: predictive accuracy, dimensional structure, and auxiliary-factor contribution. This section further examines the usability of the proposed framework in two practical mapping scenarios: full-scene inference and inter-annual prediction interpolation. Spatial statistical metrics are also introduced to validate the structural fidelity of the predictions. The six selected cities are those that reach the dual-task usability thresholds in Table 6, namely $F1 \geq 0.5$ and $R^2 \geq 0.3$: Tegucigalpa (HON), Port-au-Prince (HTI), Ouagadougou (BFA), Cairo (EGY), Karachi (PAK), and Mumbai (IND). For geographic comparison in full-scene mapping, these cities are ordered from west to east by longitude, and each city uses its optimal auxiliary configuration.

### 5.1. Full-Scene Inference and Inter-Annual Monitoring

For each city, a single TorchMLP model is trained using samples from all available GT years and then applied to the full AEF imagery for all eight years from 2017 to 2024. Each city covers approximately 100–2,000 $km^2$, with $10^6 - 2 \times 10^7$ valid pixels, as detailed in Table A3. This enables both accuracy reproduction for GT-annotated years and prediction interpolation for years without annotations.

Figure 14 reveals three main patterns. First, the spatial structure remains stable across years: for all six cities, predicted slum patches remain consistent in their core locations, without abrupt inter-annual shifts, indicating that AEF embeddings provide a stable feature basis for temporal mapping. Second, gradual expansion is observed: Ouagadougou and Cairo show slow outward expansion of slum coverage from 2017 to 2024, broadly consistent with known directions of urban growth. Third, localized annual anomalies appear: Karachi in 2022 and Mumbai in 2018 show prediction spikes, although no corresponding real morphological changes are recorded for those years. This suggests that inter-annual predictions should be interpreted together with further diagnostics of the temporal stability of AEF embeddings. Overall, the classification and density layers show visually consistent spatial patterns.

### 5.2. Supplementary Validation of Spatial Structural Fidelity

Pixel-level F1 and $R^2$ measure point-wise accuracy but do not indicate whether the model reproduces the overall spatial structure of slums. Table 7 provides supplementary validation of full-scene predictions across all GT-annotated years in the six cities using three spatial statistical metrics: SSIM, global Moran's I, and residual Moran's I. Area deviation is also included as an application-level metric. Since the GT labels are still GRAM pseudo-masks rather than independent field observations, the validation results should be interpreted as structural consistency between model predictions and GRAM labels.

The three structural metrics jointly support the structural fidelity of the full-scene predictions. In terms of morphological similarity, the mean $SSIM_{cls}$ across the six cities is 0.926, ranging from 0.894 to 0.948. Even in IND, where pixel-level F1 is relatively low, the predicted cluster shapes and boundaries remain broadly consistent with the GRAM labels. In terms of spatial clustering, $Moran_I_{Pred}$ is slightly higher than $Moran_I_{GT}$ in all cities, with differences ranging from 0.027 to 0.047. This indicates that the model preserves the spatial clustering of slums while introducing slight over-smoothing, reflecting the smoothing tendency of AEF embeddings and MLP-based prediction. Residual Moran's I ranges from 0.213 to 0.362 across the six cities and is positive in all cases, suggesting that prediction errors are not spatially random but clustered in specific subregions. This pattern is especially evident in cities with lower F1 values, such as HON and IND, indicating that future work could reduce such structured residuals through explicit spatial regularization or locally adaptive modeling.

Area deviation provides an application-level complement to the structural metrics. In HTI, PAK,

BFA, and EGY, the total slum-area error remains within 35%, suggesting that the predictions are suitable for aggregate monitoring tasks such as SDG 11.1 assessment. In contrast, HON and IND show much larger area deviations of 64.7% and 88.4%, respectively. For these two cities, full-scene predictions should be used primarily as spatial distribution references rather than as quantitative estimates of slum area.

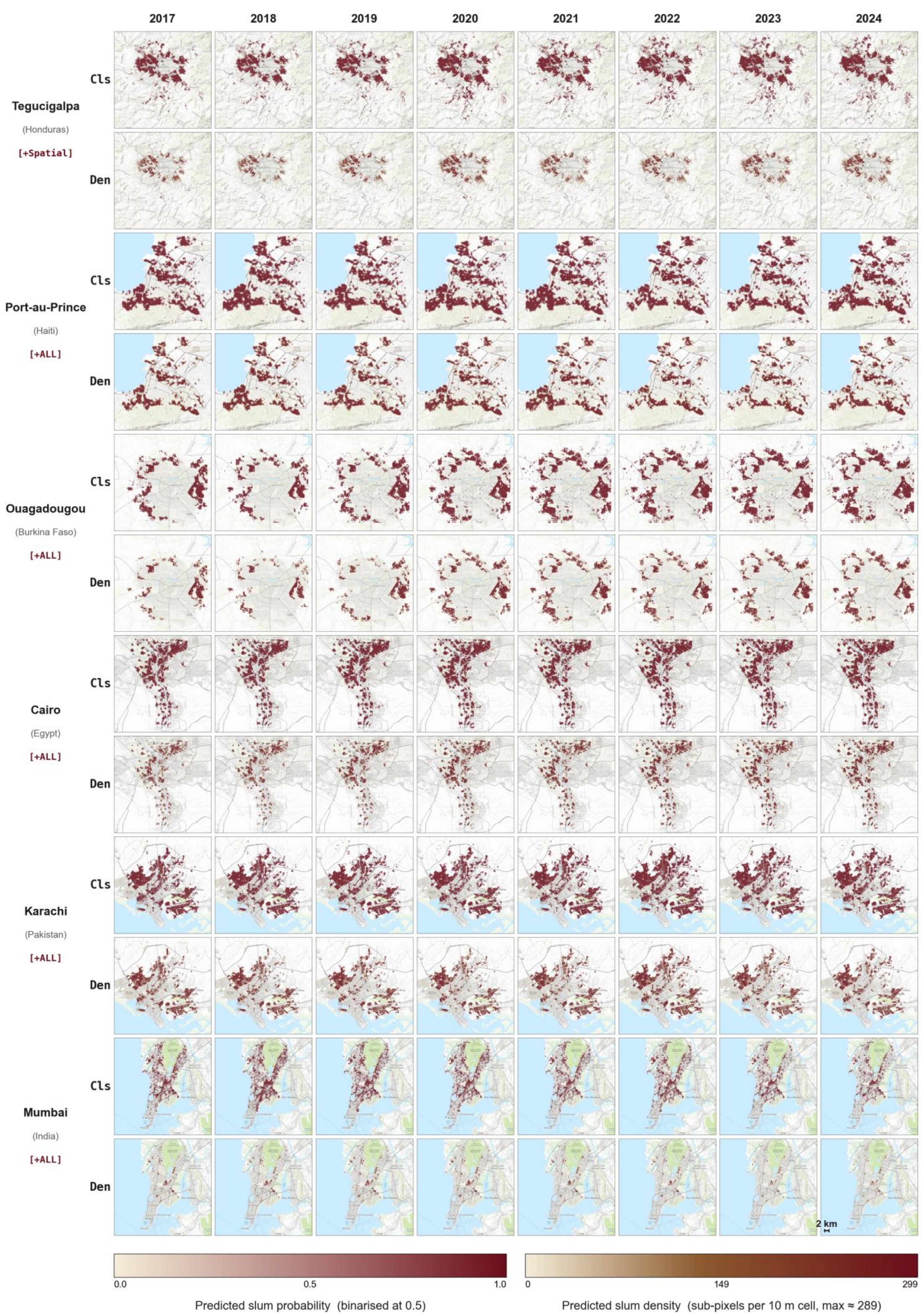

**Fig. 14 Full-AOI temporal slum mapping across six dual-task usable cities over 2017−2024, ordered west to east (inset: global locations).** Each city occupies two rows—classification (Cls,

top) and density (Den, bottom)—across eight inference years. Per city, a single TorchMLP was trained under S2 with the city's best auxiliary configuration (red brackets) and applied to the full-AOI AEF embedding for every year, including non-GT years (label imputation). Cls scale (left): probability binarised at 0.5. Den scale (right): predicted density in sub-pixels per 10 m cell (max ≈ 289). Basemap: Esri WorldTopoMap.

**Table 7. Spatial-structure validation of full-AOI predictions across the six dual-task usable cities.** F1 and IoU report classification accuracy (full-AOI evaluation; values differ from Table 6 which uses spatial CV folds). SSIM_cls quantifies cluster-shape similarity between GT and predicted maps. Global Moran's I (GT / Pred) and residual Moran's I are computed under Queen contiguity (8-neighbour, row-standardised). Area err% gives the relative deviation of predicted total slum area from GT. All metrics averaged across GT-labelled years per city; cities ordered by descending F1. Predictions: TorchMLP trained per city on all available GT years with the city's best auxiliary configuration; adaptive downsampling (4× – 8×) applied (see Appendix B).

| Code | City | F1 | IoU | SSIM_cls | Moran_GT | Moran_Pred | Residual Moran | Area err (%) |
|---|---|---|---|---|---|---|---|---|
| HTI | Port-au-Prince | 0.886 | 0.796 | 0.894 | 0.817 | 0.844 | 0.213 | 15.2 |
| PAK | Karachi | 0.879 | 0.784 | 0.948 | 0.789 | 0.825 | 0.225 | 17.6 |
| BFA | Ouagadougou | 0.847 | 0.734 | 0.944 | 0.684 | 0.717 | 0.318 | 23.3 |
| EGY | Cairo | 0.818 | 0.692 | 0.948 | 0.772 | 0.819 | 0.319 | 34.4 |
| HON | Tegucigalpa | 0.724 | 0.57 | 0.901 | 0.71 | 0.75 | 0.343 | 64.7 |
| IND | Mumbai | 0.654 | 0.486 | 0.919 | 0.603 | 0.629 | 0.362 | 88.4 |
| — | **Mean** | **0.801** | **0.677** | **0.926** | **0.728** | **0.764** | **0.297** | **40.6** |

Figure 15 provides a pixel-wise comparison of spatial clustering patterns using the four-quadrant classification of local Moran's I from LISA. Across the six cities, the core High–High slum clusters show a high degree of spatial overlap between GT and Pred, consistent with the quantitative results in Table 7, where $Moran_{I_{Pred}}$ exceeds $Moran_I_{GT}$ by only 0.027–0.047. Meanwhile, the number of High–Low and Low–High outlier patches is reduced in Pred, corresponding to the slight over-smoothing effect described above.

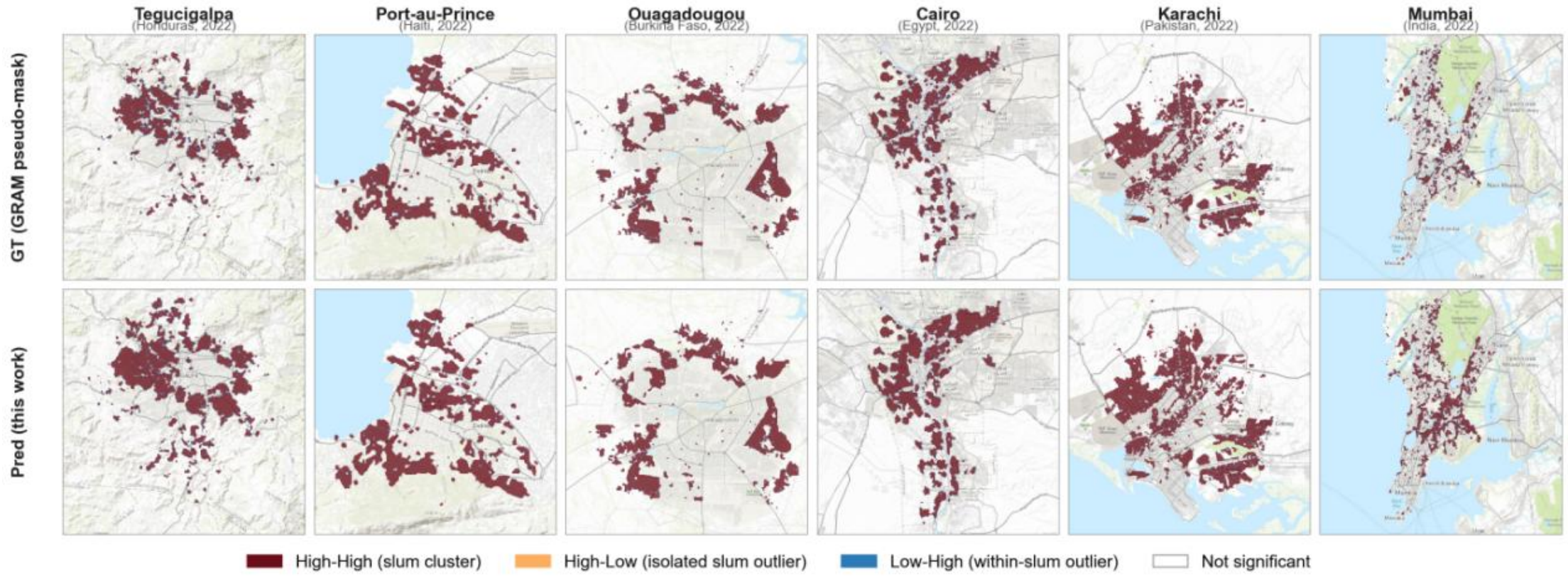

**Fig. 15 LISA local Moran's I comparison between GRAM pseudo-mask labels (top) and full-AOI predictions (bottom) across six cities at 2022, ordered west to east.** Quadrant classification: High-High (deep wine, slum cluster cores), High-Low (orange, isolated slum outliers in non-slum surrounds), Low-High (deep blue, non-slum outliers in slum clusters), Not significant (transparent, basemap visible). Computed under Queen contiguity (8-neighbour, row-standardised), 99 permutations, $p < 0.05$; adaptive downsampling (4 × – 8 × ). Predictions: TorchMLP on AEF plus the city's best auxiliary configuration. Basemap: Esri WorldTopoMap.

## 6. Discussion

### 6.1. Main Findings

With respect to the five core questions, Q1–Q5, proposed in the Introduction, this study provides the following answers.

- Q1: S2, namely cross-year training within the same city, performs best under both random and spatial evaluation. Cross-city strategies, S3 and S4, remain inferior to S2 even after sample-size alignment, indicating city-scale representation drift in AEF embeddings. The regression $R^2$is primarily driven by the discriminative ability to separate zero from non-zero pixels. The positive-pixel $R^2$, or pos $R^2$, is consistently negative across all twelve cities, with a median of $-2.00$, suggesting that AEF lacks the capacity to continuously characterize intra-pixel density gradients at the 10 m scale.
- Q2: PC36 emerges as the dominant dimension across tasks and models. Classification saturates at $k = 32$, whereas regression remains unsaturated even at $k = 64$. The secondary consensus dimensions are task-specific and non-overlapping, with PC8 and PC34 supporting classification and PC16 and PC26 supporting regression.
- Q3: For classification, C5 performs best in 11 of the 12 cities, whereas the optimal configuration for regression is distributed mainly between C4 and C5. POI contributes the largest marginal gain to density estimation ($\Delta R^2 =+ 0.064$), indicating that social-service deprivation represents a semantic signal that AEF cannot directly infer from visual texture. Five cities with atypical morphologies, namely VEN, COL, BRA, ZAF, and LKA, exceed the compensatory capacity of the current auxiliary-factor system.
- Q4–Q5: Full-scene inference shows stable spatial structures across 2017–2024, with a mean $SSIM_{cls}$ of 0.926, and the core High–High slum clusters are largely preserved. However, $Moran_I_{Pred}$is systematically higher than $Moran_I_{GT}$by 0.027–0.047, and residual Moran's I is positive in all cases, ranging from 0.213 to 0.362. These results indicate slight over-smoothing and spatially correlated prediction errors.

### 6.2. Positioning within Existing Literature

Applying AEF embeddings to slum detection and density estimation represents a first attempt in this research area. The proposed framework differs structurally from existing studies in four aspects: spatial resolution, 10 mversus VHR imagery below approximately 1.2 m; feature source, precomputed embeddings versus end-to-end image training; output type, dual classification and density tasks versus binary classification only; and evaluation design, spatial block cross-validation versus random splitting. Therefore, no directly comparable baseline can be introduced. Nevertheless, several representative studies in related directions provide useful partial references, and their relationships to this study are summarized in Table 8.

For the classification task, under random evaluation, the median F1 of S1 TorchMLP is 0.763, falling within the range of $F1 = 0.73–0.89$ reported by Wurm et al. (2019) for single-city VHR imagery in Mumbai. This indicates that publicly available 10 m embeddings can achieve a comparable level without requiring VHR imagery or transfer fine-tuning. GRAM (Lee et al., 2026), which combines VHR imagery of approximately 1.2 mwith MoE and TTA, achieves an overall mIoU of 0.859 ($F1$=0.921)across three target cities in Africa. Its accuracy is substantially higher than that of the present study, but its computational cost and imagery acquisition requirements are also much higher than those of the AEF-based route. Stark et al. (2020) reported an F1 score of up to 89% after domain adaptation in a ten-city cross-city experiment using XFCN, but this approach also relies on VHR imagery. Under spatial cross-validation, the median S2 spatial F1 of 0.616 in this study represents, to our knowledge, the first slum detection benchmark reported under a strictly spatially independent evaluation framework. Although the aforementioned VHR-based methods are not directly comparable because they primarily use random splits, this study fills an important gap in quantifying spatial generalization capacity.

For density regression, the median S2 spatial $R^2$of 0.466 represents the first systematic benchmark for pixel-level, 10 m physical slum density estimation. The $R^2$ values of 0.67–0.75 reported by Jean et al. (2016) and Yeh et al. (2020) correspond to village- or district-level wealth indices, whose spatial granularity is two to three orders of magnitude coarser. The micro-scale estimates of Chi et al. (2022) are still based on grids of approximately 2.4 km, and their target variables are aggregate socioeconomic indicators rather than physical density. Therefore, these $R^2$ values are not directly comparable with those of this study. The methodologically closest work is MOSAIKS (Rolf et al., 2021), which uses random convolutional features and linear regression to predict multiple socioeconomic variables, with $R^2$ values of 0.56–0.71. However, its spatial scale is at the census-region level, and the feature representation requires thousands of dimensions. In contrast, this study achieves pixel-level density estimation using 64-dimensional embeddings, advancing both feature efficiency and spatial granularity.

**Table 8. Positioning of this study relative to representative prior works.** Metrics are not directly comparable across rows due to differences in resolution, evaluation protocol, target variable, and spatial granularity; the table serves to delineate methodological distinctions rather than rank performance.

| Study | Resolution | Evaluation | Cities | Task | Key metric |
|---|---|---|---|---|---|
| 1(Wurm et al., 2019) | VHR (<1 m) | Random split | 1 (Mumbai) | Cls | F1 = 0.73–0.89 |
| 2(Stark et al., 2020) | VHR | Random split | 10 | Cls | F1 up to 0.89 |
| 3(Lee et al.) | ~1.2 m + TTA | Target-city inference | 12→3 | Cls | mIoU = 0.859 |
| 4(Jean et al., 2016) | Landsat 30 m | Village-level | 5 countries | Reg (wealth) | $R^2$ = 0.67–0.75 |
| 5(Yeh et al., 2020) | Landsat 30 m | District-level | ~20 k villages | Reg (wealth) | $R^2 \approx 0.70$ |
| 6(Rolf et al., 2021) | Satellite (~1 m) | Census-tract | Multi-country | Reg (multi-task) | $R^2$ = 0.56–0.71 |
| 7(Chi et al., 2022) | Multi-source | ~2.4 km grid | 56 LMICs | Reg (wealth) | $R^2$ = 0.70–0.80 |
| **This study (random)** | **AEF 10 m** | **Random 80/20** | **12** | **Cls + Reg** | **F1 = 0.763; $R^2$ = 0.624** |
| **This study (spatial)** | **AEF 10 m** | **Spatial block CV** | **12** | **Cls + Reg** | **F1 = 0.616; $R^2$ = 0.466** |

### 6.3. Limitations

This study has several limitations.

- Label-level limitations. All performance evaluations, including the spatial structure validation in section 5.2, use GRAM pseudo-masks as reference labels rather than independently collected field data. Therefore, the accuracy reported in this study reflects the consistency between two model-derived products, namely GRAM pseudo-masks aggregated from 0.6 m to 10 m and the MLP predictions based on AEF embeddings, rather than absolute accuracy relative to true slum boundaries. Any systematic biases inherent in GRAM may therefore propagate through the validation framework and remain undetected. In addition, the aggregation process from 0.6 m to 10 m introduces spatial smoothing, which compresses the contrast between high- and low-density areas. Together with the consistently negative positive-pixel $R^2$ across the twelve cities, this suggests that modeling intra-pixel density gradients at the current 10 m scale has approached the physical upper bound of the representation resolution.

- Feature-fusion limitations. In this study, auxiliary factors are appended to the 64-dimensional AEF embeddings through simple vector concatenation, without explicit interaction modeling or adaptive weighting across different semantic sources. Section 4.3 shows that C5 produces negative interactions in some cities, indicating that simple concatenation may introduce redundant or even conflicting signals. More structured fusion strategies, such as gated attention or cross-modal alignment, may further unlock the potential gains of auxiliary factors.
- Limits of city-level applicability. Five cities, namely VEN, COL, BRA, ZAF, and LKA, fail to reach the dual-task usability thresholds even under their optimal auxiliary configurations. Their very low positive-class proportions and atypical slum morphologies impose structural constraints that exceed the compensatory capacity of the current framework. In particular, the historical spatial segregation pattern in Cape Town and the extreme class imbalance in Colombo (1.4%) represent two fundamental failure modes that likely require higher-resolution imagery or more density-sensitive modeling paradigms.
- Spatial error structure. The residual Moran's I values reported in section 5.2 are positive in all six cities, ranging from 0.213 to 0.362, indicating that prediction errors are not spatially random but clustered in specific subregions. The current framework does not incorporate explicit spatial regularization. Future work could reduce such structured residuals by introducing spatial weighting constraints or locally adaptive modeling strategies.

## 7. Conclusion and Future Work

Using 69 city–year pairs from twelve cities worldwide, this study systematically evaluated the applicability of 64-dimensional AEF embeddings at 10 m resolution for the dual tasks of slum classification and sub-pixel density estimation. The main findings are as follows. S2, namely cross-year training within the same city, is the optimal strategy under both random and spatial evaluation, achieving a median spatial F1 of 0.616 and a median spatial $R^2$ of 0.466. Cross-city strategies, S3 and S4, remain inferior to S2 even after sample-size alignment, confirming the existence of city-scale representation drift in AEF embeddings. PC36 is the only dimension that remains consistently dominant across tasks and models. Classification saturates at $k = 32$, whereas regression remains unsaturated even at $k = 64$. In terms of auxiliary factors, C5 dominates across cities for classification, whereas the optimal configuration for regression is mainly distributed between C4 and C5. POI contributes the largest marginal gain to density estimation ($\Delta R^2 = +0.064$), indicating that social-service deprivation is a semantic signal that AEF visual embeddings cannot directly infer from texture alone. After six cities reached the dual-task usability thresholds, full-scene inference was conducted for 2017–2024. The mean $SSIM_{\text{cls}}$ reached 0.926, and the core slum clusters were largely preserved. These results demonstrate that publicly available 10 m embeddings combined with lightweight models can provide effective signals for slum boundary identification, although continuous characterization of intra-pixel density gradients remains constrained by the current representation resolution.

Future work can proceed in three directions. First, at the feature-fusion level, gated attention or cross-modal alignment mechanisms could replace simple vector concatenation to better exploit interactions between auxiliary factors and AEF embeddings. Second, inspired by the Mixture-of-Experts design in GRAM, city-aware routing could be incorporated into downstream models to mitigate cross-city representation drift. Alternatively, higher-resolution local imagery, such as Sentinel-2 10 m spectral bands or commercial VHR imagery, could be integrated to support density-gradient modeling and overcome the bottleneck of negative positive-pixel $R^2$. Third, to address the spatially correlated residuals identified in section 5.2, explicit spatial regularization or locally adaptive modeling approaches, such as geographically weighted regression or graph neural networks, could be introduced to reduce spatial error clustering and improve the reliability of full-scene predictions for application-level tasks such as area estimation.

## Appendix A. Study City Details

The Appendix provides supporting information for interpreting the results, including the geographic characteristics of the twelve study cities (Table A1), data coverage years (Table A2), and study-area scale and class-balance conditions (Table A3).

**Table A1. Geographic information of the 12 study cities.** Spatial Pattern provides a qualitative descriptor of slum morphology within the study-area bounding box; it is not a standardised typology. Per-year availability is given in Table A2; per-city slum-fraction ranges are given in Table A3.

| No. | Code | City | Country | Continent | Longitude | Latitude | Spatial Pattern |
|---|---|---|---|---|---|---|---|
| 1 | PAK | Karachi | Pakistan | Asia | 66.92°E–67.25°E | 24.74°N–25.10°N | High-density corridor |
| 2 | HTI | Port-au-Prince | Haiti | Latin America | 72.38°W–72.28°W | 18.50°N–18.59°N | Dense patch |
| 3 | EGY | Cairo | Egypt | Africa | 31.05°E–31.47°E | 29.72°N–30.20°N | Linear corridor |
| 4 | BFA | Ouagadougou | Burkina Faso | Africa | 1.68°W–1.41°W | 12.26°N–12.50°N | Dispersed low-density |
| 5 | HON | Tegucigalpa | Honduras | Latin America | 87.31°W–87.09°W | 13.94°N–14.16°N | Hillside terrain |
| 6 | IND | Mumbai | India | Asia | 72.77°E–72.99°E | 18.89°N–19.27°N | Vertical high-density |
| 7 | COL | Medellín | Colombia | Latin America | 75.68°W–75.50°W | 6.16°N–6.34°N | Mountain mosaic |
| 8 | KEN | Nairobi | Kenya | Africa | 36.66°E–37.10°E | 1.39°S–1.16°S | High heterogeneity |
| 9 | VEN | Caracas | Venezuela | Latin America | 67.07°W–66.86°W | 10.38°N–10.56°N | Valley patch |
| 10 | BRA | Rio de Janeiro | Brazil | Latin America | 43.80°W–43.15°W | 23.08°S–22.79°S | Dispersed (favela) |
| 11 | ZAF | Cape Town | South Africa | Africa | 18.30°E–18.88°E | 34.17°S–33.79°S | Historically segregated |
| 12 | LKA | Colombo | Sri Lanka | Asia | 79.82°E–80.06°E | 6.71°N–6.98°N | Sparse urban fringe |

**Table A2. Year-by-year data availability per city (✓ = available; — = unavailable).** Year availability is constrained by VHR imagery dates in the GRAM source dataset. The total corpus comprises 69 city – year pairs spanning 2017 – 2024. HTI is the only city with complete 8-year coverage; IND and LKA have the fewest years (4 each).

| Code | City | 2017 | 2018 | 2019 | 2020 | 2021 | 2022 | 2023 | 2024 |
|---|---|---|---|---|---|---|---|---|---|
| PAK | Karachi | ✓ | ✓ | ✓ | — | ✓ | ✓ | ✓ | ✓ |
| HTI | Port-au-Prince | ✓ | ✓ | ✓ | ✓ | ✓ | ✓ | ✓ | ✓ |
| EGY | Cairo | ✓ | — | — | ✓ | ✓ | ✓ | ✓ | ✓ |
| BFA | Ouagadougou | — | ✓ | — | ✓ | ✓ | ✓ | ✓ | ✓ |
| HON | Tegucigalpa | ✓ | — | ✓ | ✓ | — | ✓ | ✓ | ✓ |
| IND | Mumbai | ✓ | — | — | ✓ | — | ✓ | ✓ | — |

| Code | City | 2017 | 2018 | 2019 | 2020 | 2021 | 2022 | 2023 | 2024 |
|---|---|---|---|---|---|---|---|---|---|
| COL | Medellín | ✓ | ✓ | — | ✓ | ✓ | ✓ | ✓ | ✓ |
| KEN | Nairobi | ✓ | — | ✓ | ✓ | — | ✓ | ✓ | — |
| VEN | Caracas | ✓ | — | ✓ | ✓ | — | ✓ | — | ✓ |
| BRA | Rio de Janeiro | ✓ | — | ✓ | ✓ | — | ✓ | ✓ | — |
| ZAF | Cape Town | ✓ | ✓ | ✓ | — | — | ✓ | ✓ | ✓ |
| LKA | Colombo | ✓ | — | ✓ | ✓ | — | — | ✓ | — |

Specifically, Table A3 reports the AEF grid size, study-area extent, and annual ranges of slum-pixel proportions for each city, serving as a reference for understanding class-balance conditions across the study areas.

**Table A3. Per-city study-area extent and class-balance range. Total AEF Pixels = aef_w × aef_h (full study-area grid).** Slum Fraction Range = annual minimum–maximum of GRAM-labelled slum pixel proportion within the AEF bounding box across available years.

| Code | City | Total AEF Pixels | Study Area (km²) | Fraction Range |
|---|---|---|---|---|
| PAK | Karachi | 12,871,108 | 1,330 | 21.77–24.67% |
| HTI | Port-au-Prince | 1,026,596 | 106 | 18.74–26.05% |
| EGY | Cairo | 20,789,952 | 2,152 | 19.10–21.03% |
| BFA | Ouagadougou | 7,951,104 | 809 | 9.22–16.03% |
| HON | Tegucigalpa | 5,963,315 | 601 | 10.43–13.70% |
| IND | Mumbai | 9,466,041 | 979 | 7.53–8.76% |
| COL | Medellín | 3,855,600 | 389 | 5.85–7.76% |
| KEN | Nairobi | 12,215,520 | 1,232 | 5.71–6.59% |
| VEN | Caracas | 4,634,370 | 472 | 5.09–6.67% |
| BRA | Rio de Janeiro | 20,634,433 | 2,130 | 3.15–4.16% |
| ZAF | Cape Town | 21,528,342 | 2,263 | 1.04–1.83% |
| LKA | Colombo | 7,623,889 | 771 | 1.24–1.52% |